\documentclass[10pt,journal,compsoc]{IEEEtran}
\usepackage{array}
\usepackage{subfigure}
\usepackage{algorithm}
\usepackage{algorithmic}
\usepackage{multirow}
\usepackage{mathrsfs}
\usepackage{amsfonts}
\usepackage{algorithmic}
\usepackage{graphicx}
\usepackage{textcomp}
\usepackage{xcolor}

\usepackage{graphicx}
\usepackage{float}
\usepackage{subfigure}
\usepackage{algorithm}
\RequirePackage{amsmath}[2016/11/05]
\usepackage{booktabs}
\usepackage{mathrsfs}

\usepackage[justification=centering]{caption}
\ifCLASSOPTIONcompsoc
  \usepackage[nocompress]{cite}
\else
  \usepackage{cite}
\fi
\ifCLASSINFOpdf
\else
\fi
\begin{document}
%
\title{GAIN: Graph Attention \& Interaction Network \\for Inductive Semi-Supervised Learning over Large-scale Graphs}
%
%
%
%

\author{
	Yunpeng Weng,
	Xu Chen, \emph{Senior Member, IEEE},
	Liang Chen,
	Wei Liu
\IEEEcompsocitemizethanks{
	\IEEEcompsocthanksitem Yunpeng Weng, Xu Chen and Wei Liu are with the School of Data and Computer Science, Sun Yat-sen University.
	
E-mail: wengyp@mail3.sysu.edu.cn, chenxu35@mail.sysu.edu.cn, liuw79@mail3.sysu.edu.cn.
\protect\\

\IEEEcompsocthanksitem Liang Chen is with Tencent Inc. E-mail: leoncuhk@gmail.com

}
}

%
%

\markboth{}%
{Shell \MakeLowercase{\textit{et al.}}: Bare Demo of IEEEtran.cls for Computer Society Journals}
%



\IEEEtitleabstractindextext{%
\begin{abstract}
Graph Neural Networks (GNNs) have led to state-of-the-art performance on a variety of machine learning tasks such as recommendation, node classification and link prediction. Graph neural network models generate node embeddings by merging nodes features with the aggregated neighboring nodes information. Most existing GNN models exploit a single type of aggregator (e.g., mean-pooling) to aggregate neighboring nodes information, and then add or concatenate the output of aggregator to the current representation vector of the center node. However, using only a single type of aggregator is difficult to capture the different aspects of neighboring information and the simple addition or concatenation update methods limit the expressive capability of GNNs. Not only that, existing supervised or semi-supervised GNN models are trained based on the loss function of the node label, which leads to the neglect of graph structure information. In this paper, we propose a novel graph neural network architecture, Graph Attention \& Interaction Network (GAIN), for inductive learning on graphs. Unlike the previous GNN models that only utilize a single type of aggregation method, we use multiple types of aggregators to gather neighboring information in different aspects and integrate the outputs of these aggregators through the aggregator-level attention mechanism. Furthermore, we design a graph regularized loss to better capture the topological relationship of the nodes in the graph. Additionally, we first present the concept of graph feature interaction and propose a vector-wise explicit feature interaction mechanism to update the node embeddings. We conduct comprehensive experiments on two node-classification benchmarks and a real-world financial news dataset. The experiments demonstrate our GAIN model outperforms current state-of-the-art performances on all the tasks.
\end{abstract}

}

\maketitle

\IEEEdisplaynontitleabstractindextext

%
\IEEEpeerreviewmaketitle

\IEEEraisesectionheading{\section{Introduction}\label{sec:introduction}}

%
%
%
%

\IEEEPARstart{A} large number of real-world datasets are graph-structured, which contain abundant information of objects and their interactions. For instance, a social network is a typical kind of graph-structured data, the graph structure implies information such as the user's role and social relationships. In recent years, deep neural networks (DNNs) have achieved remarkable success in many tasks that tackle regular Euclidean data like images and text. However, many important tasks involve graph-structured data where regular DNNs cannot be applied directly. With the development of applications involving large-scale graph structure data, such as online social networks, applying neural networks to graph has attracted more and more attention\cite{wu2019comprehensive}. Inspired by the convolution operation with local receive fields and parameter-sharing of Convolutional Neural Networks (CNNs), several recent literatures have extended such convolution operation to the non-Euclidean graph domain\cite{henaff2015deep,duvenaud2015convolutional,kipf2016semi,hamilton2017inductive}.

According to the convolution operations defined on graphs, graph neural networks are usually categorized as spectral approaches and non-spectral approaches\cite{wu2019comprehensive}. For spectral approaches, the graph is represented with a Laplacian matrix according to the spectral theories \cite{DBLP:journals/spm/ShumanNFOV13,DBLP:journals/corr/abs-0912-3848} and the convolution operation is defined in the Fourier domain. Non-spectral approaches apply aggregation operations directly to graphs and they use aggregators to collect the neighborhood information for each node with shared weights. Nevertheless, the basic idea of these two categories of GNNs is the same and we can generalize the workflow of GNNs in two phases: 1) aggregate information for all nodes from their neighboring nodes. 2) update the vector representation of all nodes in a graph with the aggregated neighboring information. GNNs perform these two phases iteratively to learn node representation with neighborhood information in different ranges. The learned embeddings can be utilized in downstream machine learning problems such as node classification and the parameters of GNNs can be trained in the semi-supervised or fully supervised learning fashion\cite{kipf2016semi,hamilton2017inductive}.

For the aggregation phase, it is crucial to select proper aggregate operations that have the ability to adequately reflect neighborhood information and well adapt to the specific downstream machine learning task for each node. However, most existing GNN models proposed in previous literatures aggregate neighboring information with only a single type of aggregator such as mean-pooling, max-pooling and so on. The recent researches introduce a node-level multi-head attention mechanism into graph neural networks\cite{velivckovic2017graph}, but the output of each head is a linear combination of neighbor node vectors and thus graph attention neural networks still utilize only a single type of aggregator. The drawback of GNNs with a single type of aggregator is that it is difficult for them to capture neighborhood information from multiple perspectives efficiently. For instance, mean-pooling can represent the overall background information of the neighborhoods, while max-pooling can capture discriminative information. Exploiting only a single type of aggregator means the model can only capture one aspect of the neighborhood information.

For the update phase, the conventional methods for updating each target node representation vector are to average the current target node embedding $h_v$ and its aggregated neighboring information hidden vector $AGG(h_{\mathcal{N}(v)})$ or to concatenate $h_v$ with $AGG(h_{\mathcal{N}(v)})$. Nevertheless, these simple combination methods limit the expressive ability of graph neural networks since they fail to learn the feature interactions between the center node's vector with its neighborhood information hidden vector.

To address these drawbacks of existing GNNs, we propose a novel graph neural network model, Graph Attention \& Interaction Network (GAIN) in this paper. Firstly, in order to learn multifaceted neighboring information, we use different types of aggregators to deal with the neighborhood nodes information. Each type of aggregator generates a hidden vector representation of neighboring information from a specific aspect, and we further integrate the output of each aggregator into a vector through attention mechanism. With aggregator-level attention mechanism, our GAIN model has the ability to train the weights of each aggregator output according to downstream prediction tasks and make full use of neighborhood information from different perspectives. Secondly, we introduce the vector-wise explicit feature interactions into the update phase of GAIN, which aims to learn more informative representations for nodes. Feature interaction is essential for a variety of predictive problems while manual feature engineering usually comes with a high cost. DNNs are able to learn feature interactions implicitly, but they are not efficient and do not adequately cross features\cite{wang2017deep}. Explicit feature interactions have gradually become a \emph{de facto} standard in recommendation systems\cite{lian2018xdeepfm,zhang2018coupledcf} and CTR (click-through rate) prediction tasks. In this paper, we first present the concept of Graph Feature Interaction. For each node in the graph, it can be described with two feature fields:(i) its own node representation vector, (ii) the aggregated neighboring information. The graph feature interaction aims to learn the cross information between the center node's features with its neighboring information field. The expressing ability of node representation vectors can be improved via generating the vector-wise explicit feature interactions between the two feature fields for each node. For a specific example, the keyword of a post is ``graph'' and the high-frequency word contained in the comments is ``representation'', the high-frequency words of the comments of its related posts are ``network'' and ``training''. The feature interaction operation can help models more efficiently capture the combined semantics of these words, thereby helping the model determine that this post belongs to the topic of graph neural network.

In addition, we use an auto-encoder architecture to transform both the center node's current vector and aggregated neighboring information before applying explicit feature interactions. In this way, our proposed approach is enabled to learn the highly non-linear information of graph structure. According to the basic assumption of graph-structured data that close nodes in the network should be similar\cite{goyal2018graph,zhang2018network,grover2016node2vec,tang2015line}, we introduce the graph regularization into the loss function. In this way, our approach is able to better preserve the graph structure.

The GAIN model proposed in this paper is a non-spectral approach that supports mini-batch training fashion and thus our approach does not require loading the entire graph at once during the training and predicting procedure. Besides, GAIN is a general inductive model whose trained parameters can be used to generate node embeddings for previous unseen nodes like GraphSAGE\cite{hamilton2017inductive}. For large-scale graphs, GAIN is able to select a certain number of representative neighbors for each node via heuristic sampling. Therefore, our proposed GAIN is quite scalable.

In order to illustrate the effectiveness of our model, we conduct extensive experiments on several public node classification datasets, PPI, Reddit and Pubmed, and a real-world financial news dataset from Tencent, one of the largest Internet companies of China, for user response prediction. We formulate the news click prediction problem as an edge classification task on a heterogeneous network consisting of user nodes and news nodes. The experiments demonstrate that GAIN outperforms a wide variety of strong baselines and achieves state-of-the-art performance.

To summarize, the main contributions of this paper are listed as follows:
\begin{itemize}
	\item We propose a novel graph neural network model called GAIN that supports inductive learning and heuristic sampling, which makes it be competent for large-scale graph applications.
	\item We utilize multiple types of aggregators to extract neighboring information from different perspectives. The aggregator-level attention mechanism is used to determine the proper weight of each aggregator.
	\item We incorporate the auto-encoder into our models to transform both the center node's current vector and its aggregated neighboring information. We further calculate the Euclidean distance between these two transformed vectors as the graph regularization loss to preserve the graph structure information.
	\item We present an explicit graph feature interaction mechanism, which enables GAIN to learn feature crossing between each node's representation with its aggregated neighboring information and avoids extra manual feature crafting.
	\item We conduct extensive experiments on several node classification benchmarks and apply GAIN to the user response prediction problem on the real-world financial news dataset. The experiments demonstrate our model achieves state-of-the-art performance.
	
\end{itemize}
The rest of this paper is organized as follows. We discuss related work in Section 2. In section 3, we describe problem definition in detail. Section 4 introduces our proposed GAIN model in detail. Experimental explanation and results will be illustrated in Section 5. Finally, we draw our conclusion in Section 6.

\section{Related Work}
Our approach draws inspiration from several research directions including graph neural networks, network attention mechanism and explicit feature interactions. In what follows, we provide a discussion about the previous works that are related to our study.
\subsection{Graph Neural Networks (GNNs)}
Applying the neural network model on graph-structured data is a challenging problem. Graphs are non-Euclidean data, the number of neighborhood nodes for each node in the graph could be different and they are not necessary to be ordered. Gori et al. first introduced the notation of graph neural networks in the literature\cite{gori2005new} and it was further developed in \cite{micheli2009neural,scarselli2009graph}. The primitive GNNs generate the node's representation vector via node information propagation in the network with recurrent neural network architectures. They operate the propagation procedure iteratively until the node vectors achieve a stable fixed point, which requires expensive computation\cite{wu2019comprehensive}.

With the fast development of standard deep learning approaches, operating neural networks methods over graph-structured data with high effectiveness has attracted more and more attention in recent years. Inspired by the huge success of Convolutional Neural Networks (CNNs) on multiple domains, Bruna et al.\cite{bruna2013spectral} define the convolutional operation over graphs based on graph spectral theory. Since that time, the Graph Convolutional Networks (GCNs) has been developed and extended continuously. In order to better capture the spatial dependencies of nodes in graphs, the localized filter is introduced to GCN in \cite{defferrard2016convolutional} and the authors use Chebyshev polynomials to avoid computing graph Fourier bias and thus reduce the computation cost. Kipf et al. further developed GCN in \cite{kipf2016semi} by introducing first-order approximation based on ChebNet. The GCNs mentioned above are spectral-based, they achieve remarkable performance in many tasks. However, the GCNs with spectral filters require to load the whole graph at the same time and the methods mentioned above cannot handle unseen graphs.

Besides the spectral-based approaches, there are also literatures applying neural network to graphs in non-spectral ways. Non-spectral approaches define operations directly on the neighboring nodes to aggregate information for each node in the graph. In \cite{li2015gated}, Li et al. proposed a gated graph neural network (GGNN) that utilizes modern Gated Recurrent Units (GRU) in the propagation procedure. GGNN updates the node's representation via incorporating the node's own representation from the previous time step and the gathered neighboring information in this time step and computes gradients with backpropagation through time. In \cite{duvenaud2015convolutional}, the authors assign different parameter matrices to the nodes with different degrees during the aggregation procedure. The scalability of this method is limited since the number of parameters is positively related to the number of nodes with different degrees in the graph. Therefore, it cannot be applied to the tasks involving large-scale graphs. Atwood et al. proposed diffusion-convolutional neural networks (DCNNs) in \cite{atwood2016diffusion}. They used the normalized graph adjacency matrix as a transition matrix to generate information propagation in the graph. \cite{niepert2016learning} suggests that we can extract a fixed number of neighbors for each node and normalize the subgraph for each node. The normalized subgraph can be input to a convolutional neural network.

In \cite{hamilton2017inductive}, Hamilton et al. proposed a general framework, named GraphSAGE, for inductive representation learning on graphs. GraphSAGE samples a fixed number of neighbors and aggregate their information for each node. It can be applied to large scale graphs because of the mini-batch training scheme and neighbor sampling. Although different aggregators are suggested in \cite{hamilton2017inductive}, GraphSAGE can only set the same aggregator for all nodes according to manual selection. Besides, GraphSAGE adapts a uniform sampling scheme which can cause important neighbors to be ignored. To address this problem, FastGCN\cite{DBLP:conf/iclr/ChenMX18} model exploits the importance sampling scheme. FastGCN is based on Graph Convolutional Networks proposed in \cite{kipf2016semi} while they use Monte Carlo and variance reduction techniques to implement the importance neighborhood sampling and batched training scheme.

Most previous literatures demonstrate the effectiveness of GNNs with node classification tasks. GNNs also have achieved high performance on other tasks such as traffic flow forecasting\cite{zhang2018gaan,DBLP:conf/iclr/LiYS018} and recommendation systems\cite{DBLP:journals/corr/abs-1811-00855,ying2018graph,DBLP:conf/ijcai/YuYZ18}.

\subsection{Attention Mechanism}
Attention mechanism has been widely used in numerous important problems\cite{seo2016bidirectional,vaswani2017attention,wang2016attention}. Attention mechanism allows models to focus on the important part of inputs. For multi-head attention mechanism, each head calculates a attention coefficient $\alpha$ for all values in the key-value set $\{(k_1,v_1),(k_2,v_2)\cdots (k_n,v_n)\}$ when the query $q$ and key-value set are given. The coefficient is determined by each query key pair with the chosen function $f$, i.e., $\alpha_i = softmax(f(q,k_i))$. The output of each head is the linear combination of value vectors $h_v = \sum_{i=1}^{n} \alpha_iv_i$ and the final result is the combination of all heads' output.

In \cite{velivckovic2017graph}, attention mechanism was firstly introduced into GNNs and proposed graph attention networks (GATs). Intuitively, the influence of different neighbors of the node is different. If the neighbor node vectors are simply averaged in the aggregator, the information of the important neighbors might not be well utilized. To address this problem, GATs implement a node-level attention network as the aggregator. The query vector is the projected center node's feature and both key-value in each k-v pair are a neighbor node's features after projection and all head's output are concatenated into a fixed dimensional vector.

Zhang et al. proposed gated attention networks (GaAN) based on GATs in \cite{zhang2018gaan}. Different from GATs that assign all heads' output with equal weight, GaAN further computes a coefficient for each head to control its influence. In this way, the expressive ability of GNNs can be improved which makes GaAN achieve the state-of-the-art performance in several prediction tasks.

However, the previous approaches based on graph attention utilize only a single type of aggregator, which fail to capture the neighboring information from different perspectives explicitly. Although multi-head attention can help to capture different aspects of neighboring information in an implicit fashion, this approach cannot efficiently preserve different aspects of neighboring information and is lack of interpretability.

\subsection{Feature Interactions}

Feature interaction is critical for many problems. Although DNNs are able to learn feature interactions, they cross features implicitly, which leads to insufficient feature interactions. Wide \& Deep\cite{cheng2016wide}, Deep Crossing \cite{shan2016deep} and DeepFM\cite{guo2017deepfm} et al. generate embedding representation for categorical features and then learn the feature interactions.

Recently, generating feature interactions in an explicit way has been paid attention especially in recommendation problems. In \cite{wang2017deep}, Wang et al. proposed the Deep \& Cross Network (DCN) which introduced explicit feature interaction to user response (click) prediction problems. The vector representation for each sample is $X_0$ after generating embeddings for categorical features, and DCN learns feature interactions by calculating $X_0X_{i-1}^{T}W_i+b_i+X_{i-1}$ in each cross layer with trainable parameters $W$ and $b$. Lian et al. proposed the xDeepFM model in \cite{lian2018xdeepfm} with a compressed interaction network (CIN) for explicit feature interactions. Unlike DCN that achieves bit-wise level feature interactions, CIN is able to learn feature interactions at the vector-wise level. Nevertheless, the application of xDeepFM is limited due to the high time complexity of the CIN module.

The models that are able to learn feature interactions reduce the cost of manual feature engineering and improve the performance of prediction results in a variety of tasks. However, explicit feature interactions methods have not been introduced into the graph neural networks to update the node representation vectors with neighboring information.

\section{Problem Definition}
The data structures we study and notations used in this paper will be first described and then we describe the inductive semi-supervised learning problem on graphs in this section.
\begin{figure}[!t]
	\begin{center}
		\subfigure[Semi-Supervised learning for node classification. The colored nodes are labeled.] {\includegraphics[width=0.48\linewidth]{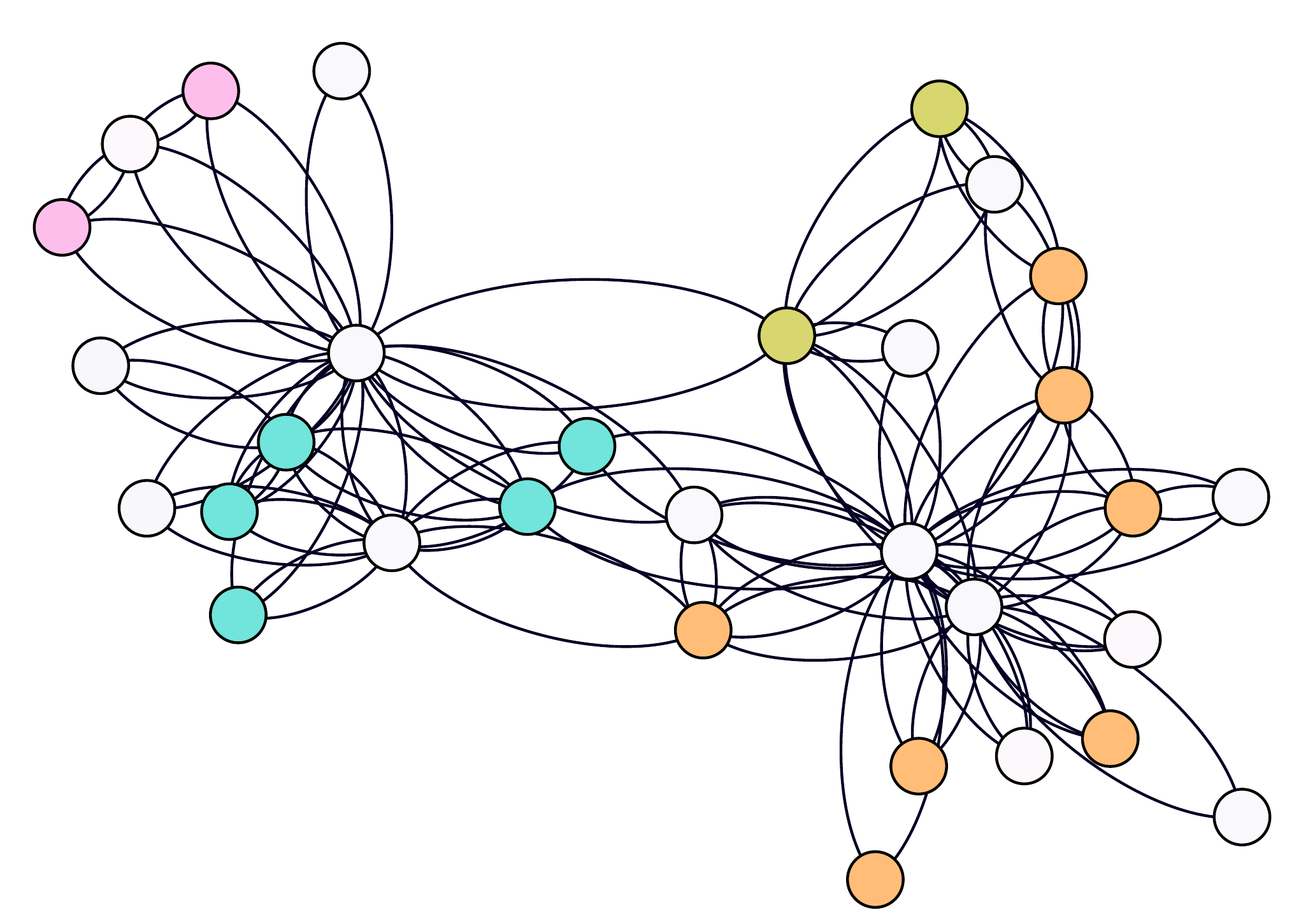}}
		\hspace*{5pt}
		\subfigure[Semi-Supervised learning for edge classification. The red and blue edges are labeled while the green edges are unlabelled.] {\includegraphics[width=0.48\linewidth]{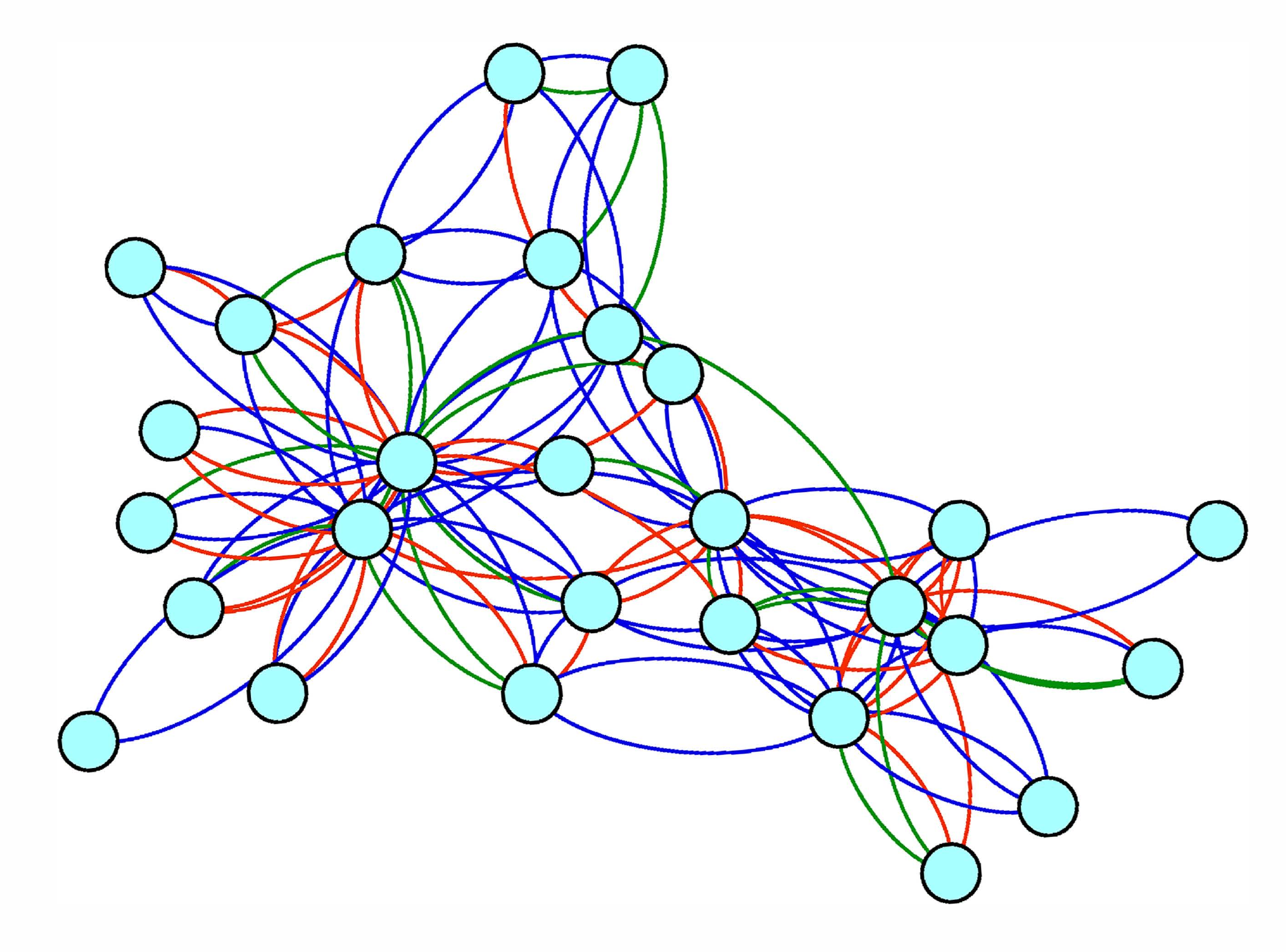}}
		\hspace*{5pt}
		\caption{The instances of semi-supervised learning on graphs}
		\label{problemins}
	\end{center}
	
\end{figure}

Let $\mathcal{G} = (\mathcal{V},\mathcal{E})$ denote the graph, where $\mathcal{V}$ is the set of nodes, and $\mathcal{E}$ is the set of edges in the graph. For each node $v$ in the graph, the representation vector of it is denoted by $h_v$ and we let $\mathcal{N}(v)$ denote the immediate neighbors of $v$.

Our proposed approach can be used for inductive (semi-)supervised learning on graphs. For each graph, there are a certain number of nodes (or edges) that are labeled and our model learns parameters according to the part of graphs with labels as shown in Fig. \ref{problemins}. The trained model can be applied to the prediction task of the unlabeled part of the graph or completely unseen graphs. In the node classification problem, we construct a subgraph for each labeled node with its neighbors from different hops and update the representation vector of the node via aggregation and update phases. Eventually, we use the updated node representation vector to predict its label. In this way, GAIN learns parameters according to the supervised loss function.

A large number of real-world application problems involving the interactions between two objects can be modeled as edge classification problem over the graph, such as advertisement click-through rate prediction, users real-life relationship prediction problems and so on. In edge classification problem, given an labeled edge $e =(v_a,v_b,y)$ where $v_a$ and $v_b$ are nodes in the edge and $y$ is the edge's label (e.g. user's response to an advertisement), the subgraphs of $v_a$ and $v_b$ are input to GAIN for aggregation and update operations, and GAIN predicts the label of edge with those two updated nodes representation vectors.

\section{Graph Attention \& Interaction Network (GAIN)}
For each node in the graph, its local neighborhood contains valuable information that can help to reveal its properties. Therefore, the representation ability of an updated center node's vector $h_v^{new}$ is mainly affected by the way of neighborhood nodes' information aggregation, and the combination way of center node's current representation $h_v$ and the aggregated neighboring information $h_{\mathcal{N}(v)}$.

\begin{figure*}[!t]
	\begin{center}
		
		\includegraphics[width=1.0\linewidth]{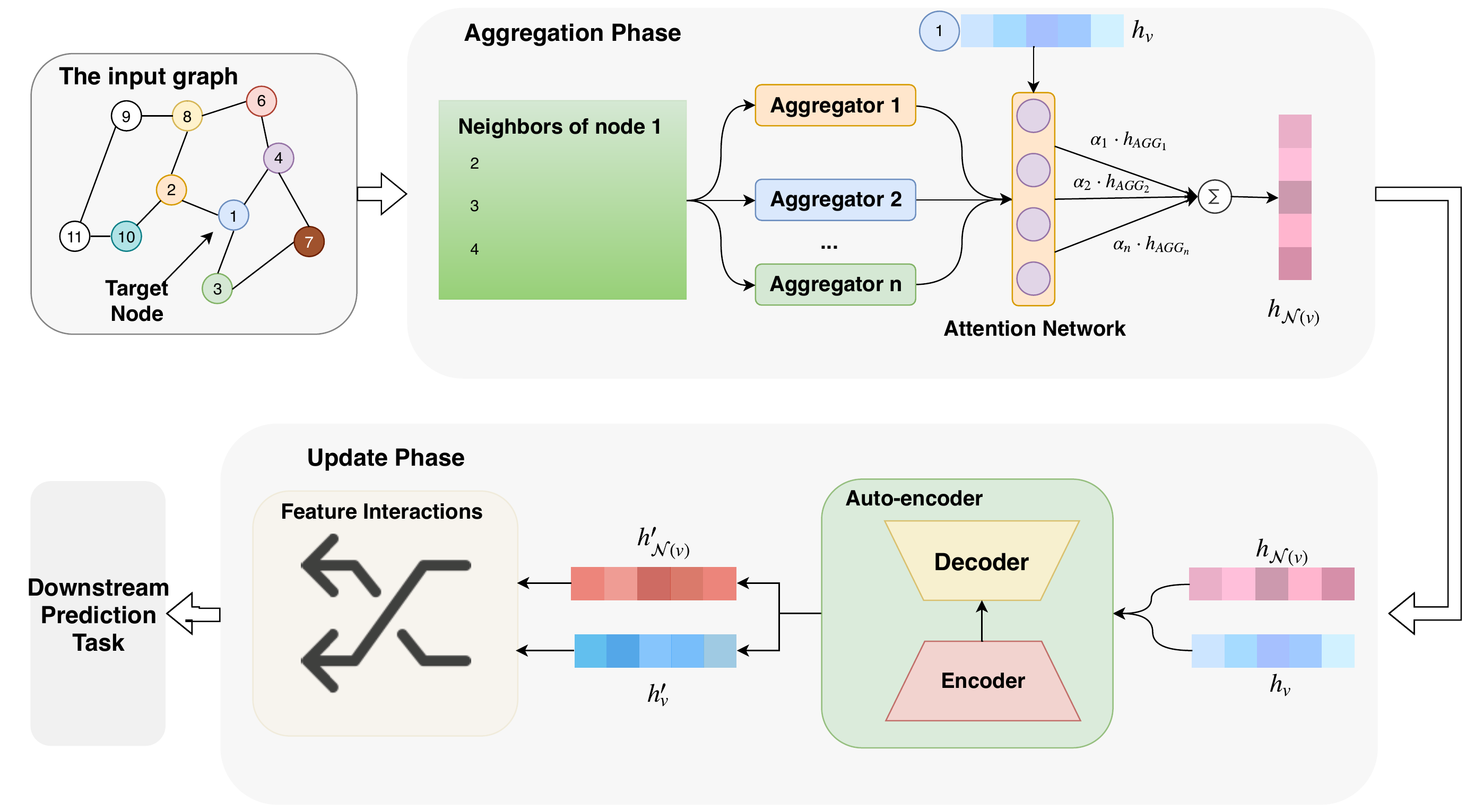}
		
		\caption{Overview of the architecture of GAIN. It consists of two critical phases: aggregation and update. For the target node $v_1$ in the input graph, GAIN aggregates its neighboring information and then transformed and cross the representation vector of $v_1$ and its neighboring information.}
		\label{overview}
	\end{center}
	
\end{figure*}
The overview of the architecture of our proposed GAIN model is shown in Fig. \ref{overview}. There are two phases in GAIN model. In the aggregation phase, we exploit several types of aggregators to generate neighboring information vectors respectively, and then integrate the output of these aggregators with the attention mechanism. In the update phase, we transform the center node's current vector $h_v$ and the output of aggregation phase $h_{N(v)}$ with the same auto-encoder, then we apply feature interactions between the transformed center node's vector and neighboring information vector to get final updated embedding for center node $v$. For a target node $v$, GAIN can aggregate the $K$-hop neighborhood information by stacking $K$ graph neural network layers. Without loss of generality, the final node representation is then projected to get the predicted result. For a multi-label classification task, we use the logistic sigmoid activation. For a multi-class classification task, we choose the softmax activation.

\subsection{Aggregation Phase: aggregator-level attention}\label{aggregation}
Different types of aggregators are able to capture the neighboring information from different perspectives explicitly. Besides, the importance of each type of aggregator might change for different downstream machine learning tasks, and their weights are not necessary to be equal. Even in the same graph, each type of aggregator may have different importance for different nodes. In our GAIN model, we use a set of aggregators $Sa = \{AGG_1,AGG_2,\cdots,AGG_n\}$ to generate a fixed dimensional neighboring information vector $AGG_i(\{h_u,\forall u \in \mathcal{N}(v) \})$.

We denote the non-linear fully connected layer as $FC_\beta = \sigma(W_\beta x+b_\beta)$ with the trainable parameters $W_\beta$ and $b_\beta$. For each center node $v$, GAIN dynamically calculates a coefficient $\alpha_{vi}$ for each aggregator with attention mechanism according to the current center node's representation $h_v$ and neighboring vectors aggregated by different types of aggregators:

\begin{equation}
\label{form1}
\begin{gathered}
e_{vi}  = {FC_{\alpha}(FC_{\theta}(h_v)\parallel FC_{\theta}(h_{AGG_i}))},\\
\alpha_{vi} = {softmax(e_{vi}) = \frac{exp(e_{vi})}{\sum_{AGG_j \in Sa} {exp(e_{vj})} }},
\end{gathered}
\end{equation}
where $\parallel$ represents concatenation operation.

For the next step, we calculate the weighted average vector of all aggregators' output with the learned attention coefficient:
\begin{equation}
\label{form2}
h_{\mathcal{N}(v)} = \sum_{AGG_i \in Sa}{\alpha_{vi} \cdot h_{AGG_i}}.
\end{equation}

\subsection{Update Phase}\label{update}
Most previous GNN variants average or concatenate $h_v$ and $h_{\mathcal{N}(v)}$ after simple linear transformation as the updated node representation. However, they might fail to capture the highly non-linear information of graph structure with such a simple update operation. Inspired by SDNE model\cite{wang2016structural} that uses auto-encoder to generate node embeddings and preserves graph structure with Laplacian eigenmaps, we first introduce an auto-encoder architecture into graph neural networks for transforming both $h_v$ and the output of aggregation phase $	h_{\mathcal{N}(v)}$. The auto-encoder module with non-linear activation function help GAIN to capture the highly non-linear information of graph structure.

\begin{figure}[h]
	\centering
	\includegraphics[width=\linewidth]{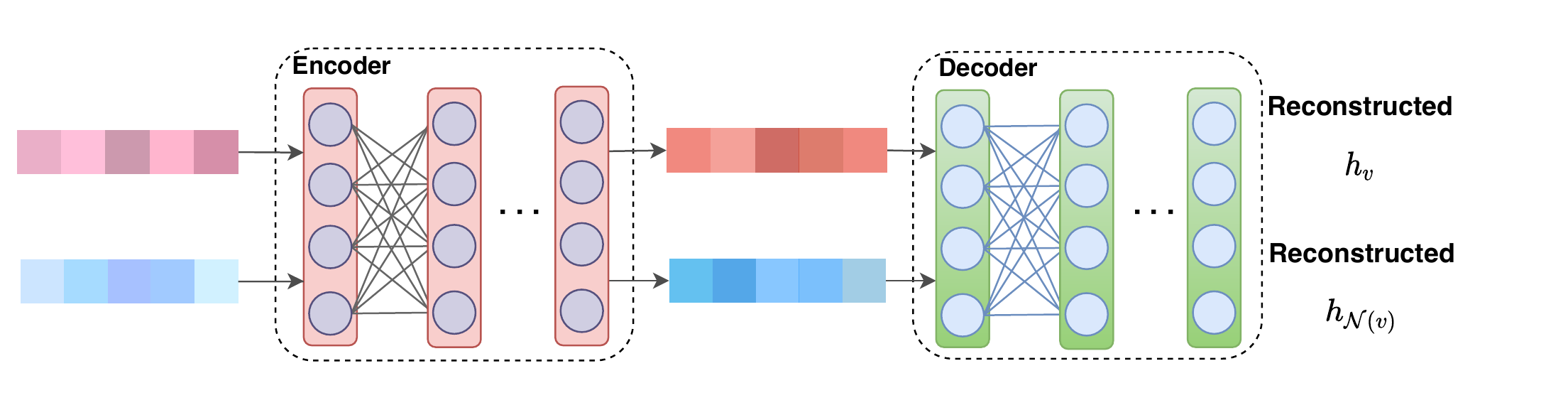}
	\caption{The auto-encoder module in GAIN.}
	\label{AE}
\end{figure}

The auto-encoder architecture used in GAIN is shown in Fig.\ref{AE}. The center node vector and the aggregated neighboring vector are transformed by the same auto-encoder and both participate in the training procedure of parameters, which can also help to learn the interactions between the center node and its neighbors implicitly. In this way, we can obtain the updated representation vector:
\begin{equation}
\begin{gathered}
h^{'}_v = Encoder(h_v),\\
h^{'}_{\mathcal{N}(v)} = Encoder(h_{\mathcal{N}(v)}).
\end{gathered}
\end{equation}
and they are then used to generate their explicit interactions in the next step.

To better capture the graph structure information, we introduce graph regularization into the loss function according to the assumption\cite{goyal2018graph} that close nodes in the graph should have similar representation:
\begin{equation}
\mathcal{L}_{greg} = \left\| h^{'}_v - h^{'}_{\mathcal{N}(v)}     \right\|_2.
\end{equation}
With the reconstructed representation vectors $\hat{h}_{v}$ and $\hat{h}_{\mathcal{N}(v)}$ after decoder operation, we can formulate the reconstruction loss for auto-encoder:
\begin{equation}
\mathcal{L}_{rec} = \left\| h_v-\hat{h}_{v} \right\|_2 + \left\| h_{\mathcal{N}(v)} - \hat{h}_{\mathcal{N}(v)} \right\|_2.
\end{equation}

\begin{figure}[h]
	\centering
	\includegraphics[width=1.0\linewidth]{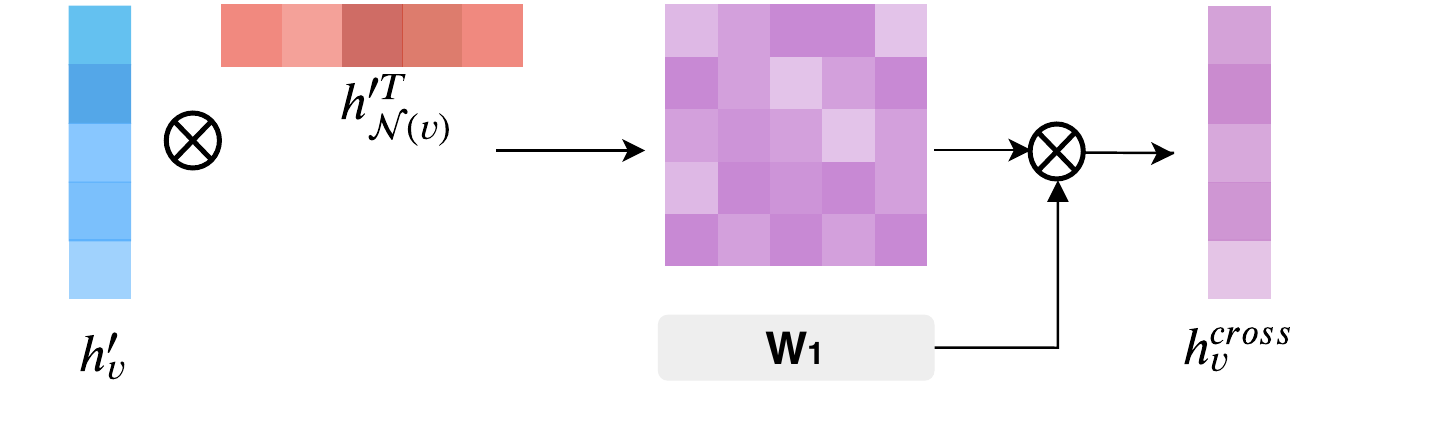}
	\caption{The feature cross operation for updating the center node vector.}

	\label{cross}
\end{figure}

In general, $h'_{v}$ and $h'_{\mathcal{N}(v)}$ could be regarded as two feature fields of the node $v$. In order to learn the feature interactions between these two fields fully and efficiently, we introduce explicit feature interaction method into our GAIN model. Inspired by the idea of Deep \& Cross Network\cite{wang2017deep}, we propose the graph feature interaction operation, which is vector-wise. The original DCN model achieves feature cross within a vector while our GAIN aims to learn the interactions between center node's vector and aggregated neighboring vector.
The cross operation between $h'_v$ and $h'_{\mathcal{N}(v)}$ is shown in Fig. \ref{cross}. The formulation of the interaction operation for $h'_v$ is:
\begin{equation}
\label{form6}
h^{cross}_{v} = h'_{v}h'^{T}_{\mathcal{N}(v)}\cdot W_1,
\end{equation}
where $h'_{v} \in \mathbb{R}^{d\times 1}$ is the current representation vector of center node $v$, $h'^{T}_{\mathcal{N}(v)} \in \mathbb{R}^{1\times d}$ is the transpose of representation vector of neighboring information and $W_1 \in \mathbb{R}^{d\times 1}$ is a trainable parameter vector. In order to reduce the space cost and computation cost, we can firstly calculate $h'_{\mathcal{N}(v)}\cdot W_1$ according to associative law. In this way, we can avoid calculating the feature map that belongs to a $d \times d$ feature space.

It should be noted that the cross operation performed between $h'_v$ and $h'_{\mathcal{N}(v)}$ is not symmetrical, and thus we need to take the similar operation to calculate $h^{corss}_{\mathcal{N}(v)}$:
\begin{equation}
\label{form7}
h^{cross}_{\mathcal{N}(v)} = h'_{\mathcal{N}(v)}h'^{T}_{v}\cdot W_2,
\end{equation}

Besides, we further use the Gated Recurrent Unit (GRU) cell to generate implicit feature interactions for the center node's vector and neighboring vector respectively.  GRU is widely used in sequence modeling with gating mechanism that uses element-wise multiplication to determine how much information from the previous hidden state can be passed to the next step. In our GAIN model, we use GRU to process the two-step sequence of ($h_v\leftarrow h_{\mathcal{N}(v)}$) to help update the representation vector of center node. The functions of $GRU(h_v,h_{\mathcal{N}(v)})$ are given as:
\begin{equation}
\label{formGRU}
\begin{gathered}
z_v = \sigma(W_zh_v+U_zh_{\mathcal{N}(v)}),\\
r_v = \sigma(W_rh_v+U_rh_{\mathcal{N}(v)}),\\
\tilde{h_v} = tanh(Wh_v +U(r_v \circ h_{\mathcal{N}(v)})),\\
h_v = (1-z_v) \circ h_{\mathcal{N}(v)}+z_v\circ\tilde{h_v},
\end{gathered}
\end{equation}
where $\circ$ represents the element-wise multiplication. For updating the representation vector of neighboring information, we can apply a similar operation for the sequence of ($ h_{\mathcal{N}(v)} \leftarrow h_v$).

The final crossed representation of center node and neighboring information can be denoted as:

\begin{equation}
\label{form8}
\begin{gathered}
h^{new}_v = h^{cross}_{v} + W_vGRU(h_v,h_{\mathcal{N}(v)}) + b_1,\\
h^{new}_{\mathcal{N}(v)} =h^{cross}_{\mathcal{N}(v)} + W_{N(v)}GRU(h_{\mathcal{N}(v)},h_{v}) + b_2,
\end{gathered}
\end{equation}
Both $h^{new}_v$ and $h^{new}_{\mathcal{N}(v)}$ have dimension of $d$ and we further concatenate them as the updated center node's representation vector for downstream prediction task.

\subsection{Model Training}
\begin{algorithm}[!h]
	\caption{GAIN minibatch forward propagation algorithm}              
	\label{alg1}                        
	\begin{algorithmic}[1]
		\REQUIRE {A batch of nodes $\mathcal{B} \in \mathcal{V}$;\\
			depth K;\\
			$\mathcal{G}=(\mathcal{V},\mathcal{E})$\\
			The set of aggregators $Sa$.
		}             
		\ENSURE The updated representation vector of each node in $\mathcal{B}$  
		\\
		/* Neighbor Sampling Procedure */
		\STATE $\mathcal{B}^{K} \gets \mathcal{B}$
		\FOR{$k$ = $K$ to 1}
		\STATE$\mathcal{B}^{(k-1)}\gets $$\mathcal{B}^k$
		\FOR{$v \in $ $\mathcal{B}^{k}$}
		\STATE $\mathcal{B}^{(k-1)}$ $\gets$ $\mathcal{B}^{(k-1)}$ $\bigcup$ NeighborsSampling($\mathcal{N}(v)$)
		\ENDFOR
		\ENDFOR
		\STATE $h_{v}\gets$ $X_v$, $\forall v \in \mathcal{B}^{0}$
		\FOR{$k$ = 1 to $K$}
		\FOR{$v \in \mathcal{B}^{k}$}
		\FOR{$AGG_i \in Sa^{k}$}
		\STATE $h_{AGG_i}^{k} \gets $  $AGG_i(\{h_u^{(k-1)},\forall u\in \mathcal{N}(v) \})  $
		\ENDFOR
		\STATE Calculate $h^{k}_{\mathcal{N}(v)}$ with aggregator-level attention mechanism, the related formulations are described in Eq. \ref{form1} and Eq.\ref{form2}.
		\STATE $h^{k'}_v \gets Encoder(h^{k}_{v}) $, $h^{k'}_{\mathcal{N}(v)} \gets Encoder(h^{k}_{\mathcal{N}(v)}) $
		\STATE Calculate $h^{new(k)}_{v}$ and $h^{new(k)}_{\mathcal{N}(v)}$ based on Eq. \ref{form6}, \ref{form7} and Eq. \ref{form8}
		
		\STATE $h^{k}_{v} \gets CONCAT(h^{new(k)}_{v},h^{new(k)}_{\mathcal{N}(v)})$
		
		\ENDFOR
		\ENDFOR
		\STATE $h_{v} \gets h^{K}_{v}$ $,\forall v \in \mathcal{B}$

		\RETURN {\{$h_v,\forall v \in \mathcal{B}$\}}
		
	\end{algorithmic}
\end{algorithm}
\begin{table*}[h] 
	\caption{Comparision of GAIN with different variants of graph neural networks}
	\centering
	\begin{tabular}{m{1.38cm}|m{6.15cm}|m{5.98cm}|m{3.2cm}} 
		
		\hline
		\label{tab0}
		
		\textbf{Model} & \textbf{Aggregation Phase} & \textbf{Update Phase} & \textbf{loss} \\ 
		\hline 
		GraphSAGE & Select a single type of aggregator such as
		lstm-aggregator to generate neighboring nodes representation $h_{\mathcal{N}_v}$.  &
		\begin{equation}\nonumber
		h_v^{k} = \sigma (W^{k}[h_v^{k-1}||h^k_{\mathcal{N}_v}])
		\end{equation}
		& Supervised loss function for (semi-)supervised training or graph-based loss function for purely unsupervised training \\
		\hline
		GCN &
		\begin{equation}\nonumber
		N_k = \tilde{D}^{-\frac{1}{2}} \tilde{A} \tilde{D}^{-\frac{1}{2}} H_{k-1}
		\end{equation}
		&
		\begin{equation}\nonumber
		H_k = \sigma (N_k W)
		\end{equation}
		& Supervised loss\\
		\hline
		GAT &
		Let $e_{vu}$ denote the attention score between center node v and its immediate neighbor u.
		\begin{equation}\nonumber
		\begin{gathered}
		e_{vu} = LeakyReLU(a[Wh_v||Wh_u])\\
		\alpha_{vu} = SOFTMAX_{u\in \mathcal{N}_v\cup u} (e_{vu})
		\\h_{N_v}^{k}= \sum_{u\in \mathcal{N}_v} \alpha_{vu} h_u^{k-1}
		\end{gathered}
		\end{equation}
		&
		\begin{equation}\nonumber
		h_v^k = \sigma(\alpha_{vv} h_v^{k-1} + h_{N_v}^k)			
		\end{equation}
		For multi-head attention mechanism with M heads, GAT repeats the aggregation and update phases M times with different parameters and average or concatenate the M heads' outputs.
		& Supervised loss\\
		\hline
		GaAN & Use attention network to aggregate the center node's local neighborhood information like GAT.& Weighted sum of center node vector and neighboring vector. Calculate an additional gate to control the importance of each head when multi-head attention mechanism is applied.
		
		& Supervised loss\\
		\hline
		
		GAIN(ours) & Aggregate neighborhood information through the aggregator-level attention mechanism with multiple types of aggregators.
		&
		The auto-encoder transformation procedure and explicit graph feature interactions, which has been described in \ref{update}.
		
		&  Supervised loss + reconstruction loss + graph regularization\\
		\hline
	\end{tabular}
\end{table*}

The GAIN model proposed in this paper can be trained in an inductive supervised or semi-supervised fashion over graphs. Different from most existing spectral-based GNNs, our model doesn't have to be retrained for unseen graphs since what GAIN learns are the parameters that can be shared and transferred.

In order to enhance the scalability for large-scale graphs, GAIN supports neighbor sampling before training. For many complex networks such as online social networks, the nodes on the graph can have thousands of neighbors, which leads to the high memory cost for model training. GraphSAGE\cite{hamilton2017inductive} addresses this problem with uniform sampling scheme. However, sampling a fixed number of neighbors for each node randomly cannot guarantee that the sampled neighbor nodes are representative enough.

To address this problem, we exploit a heuristic score function like Jaccard Coefficient (JC) to measure the importance of neighbors for sampling when the node's neighbor size is larger than the sample size. In this way, the neighbors that have similar topological structure to the center node in the graph will have greater probability of being sampled. For a center node $v$, the sampling probability of its each neighbor node $i$ could be defined as:
\begin{equation}
\begin{gathered}
P(i|v) = \frac{JC(v,i)\cdot w_{vi}+\epsilon}{\sum_{j\in \mathcal{N}(v)} {JC(v,j)\cdot w_{vj}+\epsilon}},\\
JC(v,i) = \frac{|\mathcal{N}(v) \bigcap \mathcal{N}(i)|}{|\mathcal{N}(v) \bigcup \mathcal{N}(i)|},
\end{gathered}
\end{equation}
where $\epsilon$ is a hyperparameter which is used to achieve smoothing effect and $w_{vi}$ is the weight of edge between node $v$ and $i$. For each center node, it incurs a time complexity of $O(d)$ to calculate the Jaccard Coefficient and sampling probability where the $d$ is the average degree of the graph. Besides Jaccard Coefficient, our proposed GAIN also supports calculating the sampling probability with other heuristic scores including Common Neighbors and Node Degree. We can select a suitable heuristic score according to the specific task, aggregator types and dataset.

We summarize the forward propagation of GAIN model for generating embeddings for the nodes contained in the minibatch in Algorithm \ref{alg1}. The procedure of GAIN model includes heuristic neighborhood sampling, neighborhood information aggregation and node embedding update.

The updated representation vector $h_{v}$ is then used for downstream prediction tasks, and the parameters of GAIN are trained with the loss function in the backpropagation procedure. The total loss is denoted as:
\begin{equation}
\begin{split}
\mathcal{L} =& \mathcal{L}_{sup} + \lambda_1\mathcal{L}_{greg} + \lambda_2\mathcal{L}_{rec}\\
&= \frac{1}{M}\sum_{i}\{\frac{1}{|\mathcal{Y}|}{\sum_{k\in \mathcal{Y}} loss(y_k^i,\hat{y_k^i})} \\
&+\lambda_1 \left\| h^{'}_{v_i} - h^{'}_{\mathcal{N}(v_i)}     \right\|_2  \\
&+\lambda_2\left\| h_{v_i}-\hat{h}_{v_i} \right\|_2 + \lambda_2 \left\| h_{\mathcal{N}(v_i)} - \hat{h}_{\mathcal{N}(v_i)} \right\|_2\},
\end{split}
\end{equation}
where M is the number of labeled nodes (edges), $\mathcal{L}_{sup}$ is the supervised loss involving with the labeled part of graphs, $\mathcal{L}_{rec}$ is the reconstruction loss of auto-encoder and $\mathcal{L}_{greg}$ is the graph regularization loss for preserving graph structure information. $\lambda_1$ and $\lambda_2$ are hyperparameters.

\subsection{Comparision with Other GNNs}
The GAIN model proposed in this paper is quite different from the various variants of the graph neural network proposed in previous literatures. To be specific, the aggregation phase, update phase and loss functions of our GAIN model are significantly different from those of other graph neural network models. The comparision of GAIN and other GNNs is summarized in Table \ref{tab0}. Due to the adoption of heuristic sampling strategy and aggregator-level attention mechanism, the GAIN model can efficiently and comprehensively learn the information of neighbor nodes. Compared with GNNs using random sampling strategy and a single type of aggregator, the GAIN model has better performance and stability.

\section{Experiments}
We conduct comprehensive experiments to demonstrate the effectiveness of our proposed GAIN model. We perform comparative evaluations of GAIN with the state-of-the-art models on two tasks: (i) node classification on two widely used benchmark datasets: PPI and Reddit\cite{hamilton2017inductive}, (ii) user response prediction problem on real-world financial news dataset from Tencent. Besides, we conduct additional node classification experiments on a relatively small dataset, Pubmed dataset, with the same training/validation/test split setting as the work of FastGCN\cite{DBLP:conf/iclr/ChenMX18}. For all experiments, we run GAIN model multiple times with different random seeds and report the average test performances with the standard deviation.
\subsection{Data Description}\label{data}

We first introduce the datasets used in experiments. Most previous work related to graph neural networks only performed experiments on node classification problem. In this paper, we not only perform node classification tasks on the public benchmark datasets, but also prove the applicability of our proposed GAIN model through the user response prediction task.
\begin{itemize}
	\item \textbf{PPI}. The Protein-Protein Interaction (PPI) dataset contains 24 biological graphs where each node represents a protein and each edge represents the interactions between proteins. Each graph in PPI corresponds to a human tissue\cite{zitnik2017predicting}.  Each node in the PPI dataset is labeled with the biological functions of the protein, the number of classes is 121 and a node could possess multiple different labels. There are 20 subgraphs in the training set, two in the validation set and the rest two subgraphs in the testing set.
	
	\item \textbf{Reddit}. Reddit is an online discussion forum. Users can post posts on different topics on Reddit. In the graph of  Reddit dataset, each node represents a post and there is an edge between two different nodes if they are commented by the same user. The nodes are labeled according to the community these posts belong to.
	
	PPI and Reddit have been widely used as the node classification benchmark datasets. The statistics of the datasets are summarized in Table. \ref{tab1}.
	
	\begin{table}
		
		\caption{The PPI and Reddit datasets for inductive node classification problems.}
		\label{tab1}
		\begin{tabular}{ccccl}
			\toprule
			Dataset& \#Nodes&\#Edges&\#Features(Node)&\#Classes\\
			\midrule
			PPI & 56,944 & 818,716 & 50 &121(multilabel) \\
			Reddit & 232,965& 11,606,919 & 602& 41(single) \\
			
			\bottomrule
		\end{tabular}
	\end{table}
	\item \textbf{Pubmed}. Pubmed is a citation network. It contains 19717 nodes, 44338 edges. Each node in the graph is a document with single label, the number of classes is 3. Specifically, we use the same train/validation/test splits as the work of FastGCN with 18,217 training nodes, 500 validation nodes and 1,000 testing nodes.
	\item \textbf{Tencent Financial News}. Tencent is one of the largest Internet companies in China. We collect 1,122,976 users response records from the historical logs from March 25, 2019 to April 14, 2019. In this problem, we aim to predict whether users will click the financial news that is explored to them or not. We formulate the problem as an edge classification problem over a heterogeneous graph. Unlike the other datasets used in this paper, the graph of this dataset contains more than one type of nodes, but is composed of user nodes and news nodes. For features, we trained the word2vec model\cite{mikolov2013efficient,mikolov2013distributed} to generate a 50-dimensional word vector for all words of news in the dataset. For each news, we concatenate(i) the average word vector of the news title,(ii) the average word vector of the news content. For each user, we calculate the average feature vector of the news that the user has clicked within 15 days before March 25, 2019. The training set contains the records from March 25 to April 11, the validation set contains the records from April 12 to 12:00 of April 13 and the samples in the remaining 36 hours are used for testing.
\end{itemize}

\begin{figure}[t]
	\centering
	\includegraphics[width=0.8\linewidth]{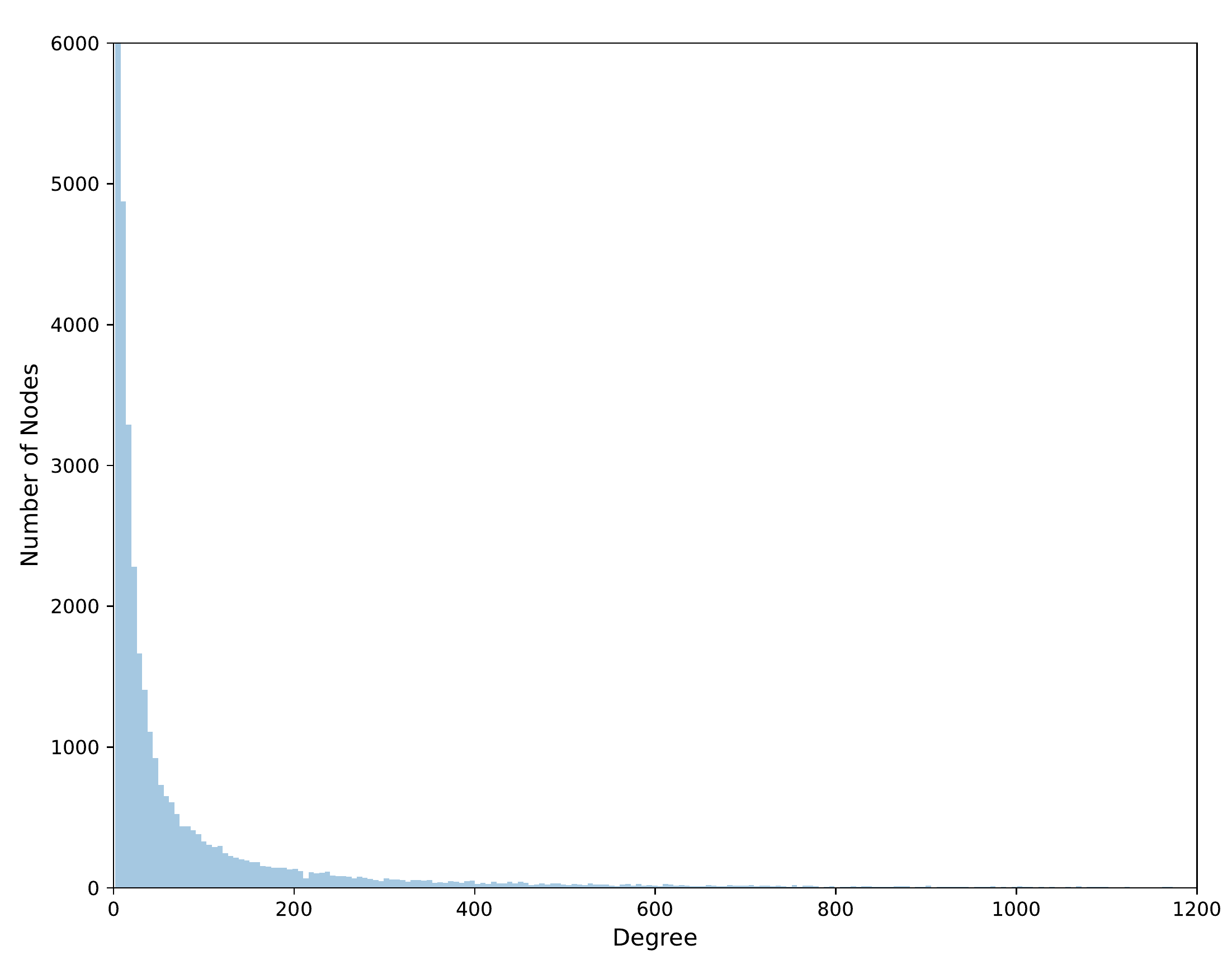}
	\caption{The node degree distribution of User-news Response Network.}
	\label{vis2}
\end{figure}

The User-News network is a bipartite graph, the 1-hop neighbors of each user node are news nodes and 1-hop neighbors of each news node are users. As shown in Fig. \ref{vis2}, most nodes in the User-News heterogeneous graph have a small number of neighbors and a tiny minority of nodes have a great number of neighbors, which indicates the long-tailed nature of degree distribution of the graph. On one hand, we can choose a relative small sample size to remain all neighbors of most nodes.On the other hand, it's significant but challenging to sample the most representative neighbor nodes for the nodes with a larger number of neighbors.

\subsection{Experimental Set-up}
The aggregators used in GAIN model are described as follows:
\begin{itemize}
	\item Average Pooling: we take the element-wise average of $\{h_{i}, \forall i \in \mathcal{N}{(v)}\}$to calculate the representation of neighboring information.
	\item Max Pooling: it's similar to average pooling but we take the element-wise max operation on neighborhood nodes vectors.
	\item Importance Pooling: The neighbors of each node are ordered according to the sampling probability before being inputted to GAIN model. We further train a gate parameter to scale each neighbor node's vector and then calculate the weighted average of all neighbor nodes vector for each center node.
\end{itemize}
For all datasets used in experiments, the number of layers is set to be 2 for GAIN model. We choose Adam optimizer\cite{kingma2014adam} to train our model. Dropout rate is set to 0 and we take L2 normalization on all updated node embeddings. We use $S_i$ to denote the maximum number of sampled neighbors in the $i-th$ sampling steps.

In the PPI experiments, the dimension $d$ is set to be 512 for each GAIN layer. The learning rate is fixed to 0.0025. The graph regularization strength $\lambda_1$ is set to be 0.004 and reconstruction loss strength of auto-encoder $\lambda_2$ is set to be 0.05 with a grid search over $\lambda_1 \in \{0,0.001,0.002,0.003,0.004,0.005\}$ and $\lambda_2 \in \{0.01,0.05,0.10,0.15,0.20\}$. The maximum number of sampled neighbors are both 220 at depth $K = 1$ and $K = 2$. We adopt heuristic sampling according to Jaccard Coefficient. We apply sigmoid activation on the final representation of nodes for the multi-label node classification.

For Reddit dataset, the learning rate is initialized to 0.0002 and gradually decreases to 0.00005 with the decay rate of 0.5 once the micro-F1 score on the validation set has not increased for a epoch.
The dimension $d$ is set to be 512 in the first layer of GAIN and 1024 in the second GAIN layer. In the neighbor node sampling process, we use the node degree as the heuristic score, and preferentially preserve the neighbors with large degree. The graph regularization strength $\lambda_1$ is set to 0.0004 and reconstruction loss strength of auto-encoder is set to 0.03 with a grid search over $\lambda_1 \in \{0,0.0001,0.0002,0.0003,0.0004,0.0005\}$ and $\lambda_2 \in \{0.01,0.02,0.03,0.04,0.05\}$. With the obtained node embedding, we apply softmax layer to get the predicted result for this multi-class node classification task.

For Pubmed dataset, the learning rate is initialized to 0.0025 and decreases as the setting for Reddit. The dimension $d$ is set to be 512 and 128 for the first and the second layer. $\lambda_1$ is 0.008 and $\lambda_2$ is 0.4. The sampling neighborhood sizes are 70 and 50 at the depth $ K = 1 $ and $ K = 2$.

In the user response to financial news prediction experiments, $d$ is set to be 128 in the first GAIN layer. In the second layer, it's set to be 512. The learning rate is fixed to 0.0002. For all GNN models compared in this experiment, we perform the neighborhood sampling and set both $S_1$ and $S_2$ to 70. For GAIN model, we use Jaccard Coefficient as the heuristic score to calculate the neighbors sampling probability. In this experiment, $\lambda_1$ is set to 0.004 and $\lambda_2$ is set to 0.05 according the grid search on $\lambda_1 \in \{0,0.002,0.004,0.006,0.008,0.01\}$ and $\lambda_2 \in \{0.01,0.02,0.03,0.04,0.05\}$. For all GNN models compared in this experiment, the updated news representation and user representation are then concatenated, and this concatenated vector is fed through 3 fully connected layers to get the output.
\begin{table}
	\caption{The experimental test results (micro-F1 scores) for the inductive node classification tasks. }
	\label{tab2}
	\scalebox{1.25}{
		\begin{tabular}{ccl}
			\toprule
			Models / Datasets& PPI& Reddit\\
			\midrule
			GraphSAGE\cite{hamilton2017inductive} & 61.2 & 95.4\\
			DGI\cite{DBLP:journals/corr/abs-1809-10341}(unsupervised) & 63.8 $\pm$ 0.2 & 94.0$\pm$0.1\\
			GraphSAGE$^{*}$  &  86.06$\pm$0.08& 96.35$\pm$0.02 \\
			GraphSAGE$^{*}$(HS)  & 86.90$\pm$0.06& 96.35$\pm$0.03 \\
			
			GAT\cite{velivckovic2017graph} &97.3 $\pm$ 0.2 & -  \\
			FastGCN\cite{DBLP:conf/iclr/ChenMX18} & - & 93.7  \\
			GeniePath\cite{DBLP:journals/corr/abs-1802-00910} & 98.5 & -  \\
			
			Cluster-GCN\cite{DBLP:conf/kdd/ChiangLSLBH19} & 99.36 &  96.60 \\
			
			GaAN\cite{zhang2018gaan} & 98.71 $\pm$ 0.02 & 96.83$\pm$0.03\\
			\textbf{GAIN(ours)}& \textbf{99.37 $\pm$ 0.01} & \textbf{97.03$\pm$0.03} \\
			
			\bottomrule
		\end{tabular}
	}
\end{table}
\begin{center}
	\begin{table}

		\caption{Summary of results in terms of micro-F1 score for Pubmed}
		\centering
		\label{tab_pubmed}
		\scalebox{1.25}{
			\begin{tabular}{ccl}
				\toprule
				Models &  Micro-F1\\
				\midrule
				GraphSAGE-GCN & 84.9 \\
				GraphSAGE-MEAN & 88.8 \\
				GCN (batched)  & 86.7 \\
				GCN (original)  & 87.5 \\
				FastGCN &  88.0 \\
				\textbf{GAIN(ours)}& \textbf{90.8 $\pm$ 0.4} \\
				
				\bottomrule
			\end{tabular}
		}
	\end{table}
\end{center}
\begin{table*}[!h]
	\caption{Results of the test micro-F1 score on the Reddit dataset with different sampling neighborhood sizes at the first and second depth of GNNs. For the GNNs with multi-head aggregator, the number of attention head is denoted as K.}
	\label{sizes}
	\centering
	\scalebox{1.23}{
		\begin{tabular}{ccccc}
			\toprule
			\centering
			Models/Sample Sizes &$(S_1=25,S_2=10)$ & $(S_1=50,S_2=20)$& $(S_1=100,S_2=40)$&$(S_1=200,S_2=80)$\\
			\midrule
			
			Avg. pooling\cite{zhang2018gaan} & 95.78$\pm$ 0.07& 96.11$\pm$ 0.07& 96.28$\pm$ 0.05 &96.35$\pm$0.02 \\
			
			Max pooling\cite{zhang2018gaan} & 95.62$\pm$ 0.03& 96.06$\pm$ 0.09& 96.18$\pm$ 0.11 &96.33$\pm$0.04 \\
			
			Attention (K=1)\cite{zhang2018gaan} & 96.15$\pm$ 0.06& 96.40$\pm$ 0.05& 96.48$\pm$ 0.02 &96.54$\pm$0.07 \\
			
			Attention (K=2)\cite{zhang2018gaan} & 96.19$\pm$ 0.07& 96.40$\pm$ 0.04& 96.52$\pm$ 0.02 &96.57$\pm$0.02 \\
			Attention (K=4)\cite{zhang2018gaan} & 96.11$\pm$ 0.06& 96.40$\pm$ 0.02& 96.49$\pm$ 0.03 &96.56$\pm$0.02 \\
			Attention (K=8)\cite{zhang2018gaan} & 96.10$\pm$ 0.03& 96.38$\pm$ 0.01& 96.50$\pm$ 0.04 &96.53$\pm$0.02 \\
			GaAN (K=1)\cite{zhang2018gaan} & 96.29$\pm$ 0.05& 96.50$\pm$ 0.08& 96.67$\pm$ 0.04 &96.73$\pm$0.05 \\
			GaAN (K=2)\cite{zhang2018gaan} & 96.33$\pm$ 0.02& 96.59$\pm$ 0.02& 96.71$\pm$ 0.05 &96.82$\pm$0.05 \\	
			GaAN (K=4)\cite{zhang2018gaan} & 96.36$\pm$ 0.03& 96.60$\pm$ 0.03& 96.73$\pm$ 0.04 &96.83$\pm$0.03 \\	
			GaAN (K=8)\cite{zhang2018gaan} & 96.31$\pm$ 0.13& 96.60$\pm$ 0.02& 96.75$\pm$ 0.03 &96.79$\pm$0.08 \\	
			\textbf{GAIN (ours)} & \textbf{96.67$\pm$0.04}& \textbf{96.87$\pm$ 0.02}& \textbf{96.92$\pm$ 0.02} &\textbf{97.03$\pm$0.03} \\	
			
			\bottomrule
			
		\end{tabular}
	}
\end{table*}
\subsection{Inductive Node Classification}
The experiments on inductive node classification task are conducted on public benchmark datasets. We compare our proposed GAIN models with several strong baselines including the state-of-the-art models for these datasets.
The experimental results are evaluated by micro-averaged F1 score as in most previous papers on these datasets.
The results of PPI and Reddit are reported in Table \ref{tab2}. Note that the experimental results reported in \cite{hamilton2017inductive} are not the best performance that GraphSAGE model can achieve, because the authors used only a small number of neighbor nodes for each center node and didn't train the model to converge. Therefore, for a fair comparison, we also report the best performance of GraphSAGE we obtained with the same neighbors sampling size and the hidden dimension of each layer as GAIN model for PPI dataset. For Reddit dataset, we also trained a GraphSAGE model with proper hyperparameter tuning. To be specific, we set the dimensionality of aggregators to be 512 and 1024 for the first layer and second layer respectively and use the concatenation operation to integrate the aggregated neighboring information vector with the center node's embedding. The learning rate is set to the same as GAIN. Besides, the neighborhood sample sizes $S_1 = 200, S_2 = 80$ are the same as GaAN and our GAIN model for Reddit dataset.  GraphSAGE$^{*} $ in the Table \ref{tab2} corresponds to the best GraphSAGE result we obtained. We further provide the experimental results of applying heuristic sampling strategy as GAIN to the GraphSAGE model, which is denoted as GraphSAGE$^{*}$(HS).

The results shown in Table \ref{tab2} demonstrate that our proposed GAIN model outperforms the current state-of-the-art performance on these two public benchmark datasets. The best performance reported in previous papers are included. Note that although Cluster-GCN achieves state-of-the-art performance on PPI dataset, they train a 5-layer GCN with 2048 hidden units while our GAIN only utilizes 2 layers.  With aggregator-level attention mechanism, our model can capture  different aspects of neighboring information and the developed updating phase with graph regularization and an explicit feature interaction improves the expressive ability of node embeddings. In addition, we report the testing results on Pubmed dataset in the Table \ref{tab_pubmed} and compare with the results reported in FastGCN\cite{DBLP:conf/iclr/ChenMX18}. It shows that GAIN model outperforms other comparison methods in this dataset.

In order to further illustrate the effectiveness of our proposed GAIN model, we conduct experiments on Reddit dataset with different neighborhood sample sizes and compare with the state-of-the-art performance of several graph neural network architectures reported in \cite{zhang2018gaan}. As shown in Table \ref{sizes}, our proposed GAIN model consistently outperforms the compared GNNs and achieves state-of-the-art performance with different sample sizes. Furthermore, with the aggregator-level attention and explicit feature interactions, GAIN could achieve high performance with small neighborhood sample size, which indicates that GAIN is able to efficiently exploit the neighboring information. We can also find that in the reddit dataset, sampling more neighbors for each node in the Reddit graph usually leads to an increase in prediction performance.

\subsection{User Response Prediction}
\begin{table}

	\caption{The experimental results for user response prediction task.}
	
	\label{news}
	\centering
	\scalebox{1.21}{
		\begin{tabular}{ccl}
			\toprule
			Models / Metrics& AUC(\%)& Logloss\\
			\midrule
			GraphSAGE-MEAN & 71.33$\pm$0.1011 & 0.4526$\pm$0.0165\\
			GraphSAGE-MAX & 68.34$\pm$0.3253 & 0.4640$\pm$0.0035\\
			GraphSAGE-LSTM & 67.30$\pm$0.1710 & 0.4741$\pm$0.0171\\
			GraphSAGE-GCN & 66.72$\pm$0.4274 & 0.4600$\pm$0.0052\\
			GAT & 70.52$\pm$0.5625 & 0.4521$\pm$0.0172\\
			
			FM & 60.24$\pm$ 0.0760 & 0.5065$\pm$0.0008\\
			DNN & 65.09$\pm$ 0.1297 & 0.4894$\pm$0.0010\\
			DeepFM & 65.33$\pm$ 0.0052 & 0.4850$\pm$0.0050\\		
			xDeepFM & 65.97$\pm$0.0010 & 0.4808$\pm$0.0031\\
			DCN & 67.08$\pm$0.0003 & 0.4911$\pm$0.0016\\	
			
			\textbf{GAIN(ours)} & \textbf{74.15$\pm$0.0351} & \textbf{0.4250$\pm$0.0004}\\	
			
			\bottomrule
		\end{tabular}
	}
\end{table}
\begin{figure}[thb]
	\centering
	\includegraphics[width=0.9\linewidth]{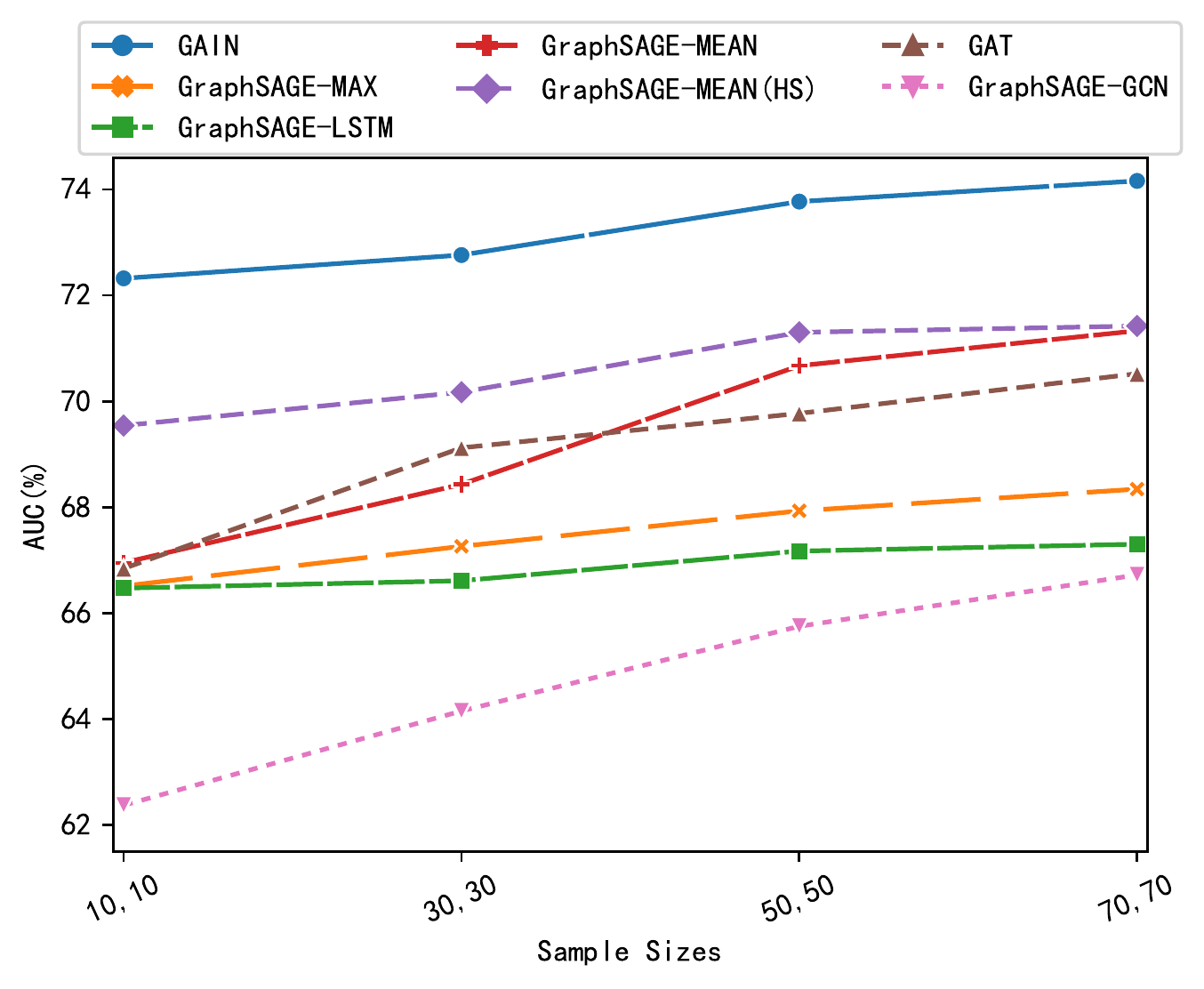}
	\caption{Results of the test AUC on the Tencent Financial News dataset with different sampling neighborhood sizes at the first and second depth of GNNs. }
	
	\label{sizes_fig}
\end{figure}

\begin{figure*}[h]
	
	\begin{center}
		
		\subfigure[Testing Micro-F1 on PPI dataset.] {\includegraphics[width=0.32\linewidth]{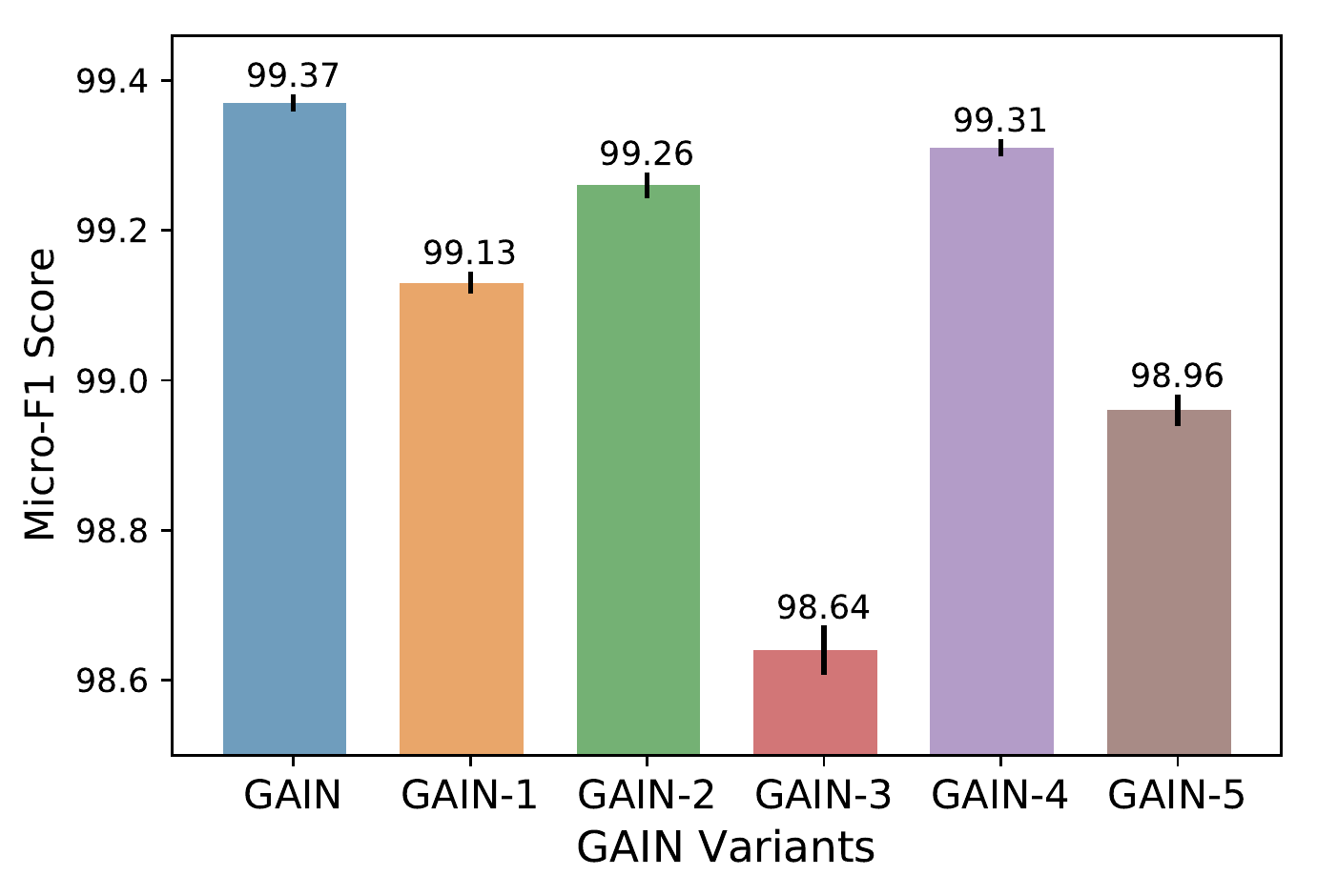}}
		\hspace*{3pt}
		\subfigure[Testing Micro-F1 on Reddit dataset.] {\includegraphics[width=0.32\linewidth]{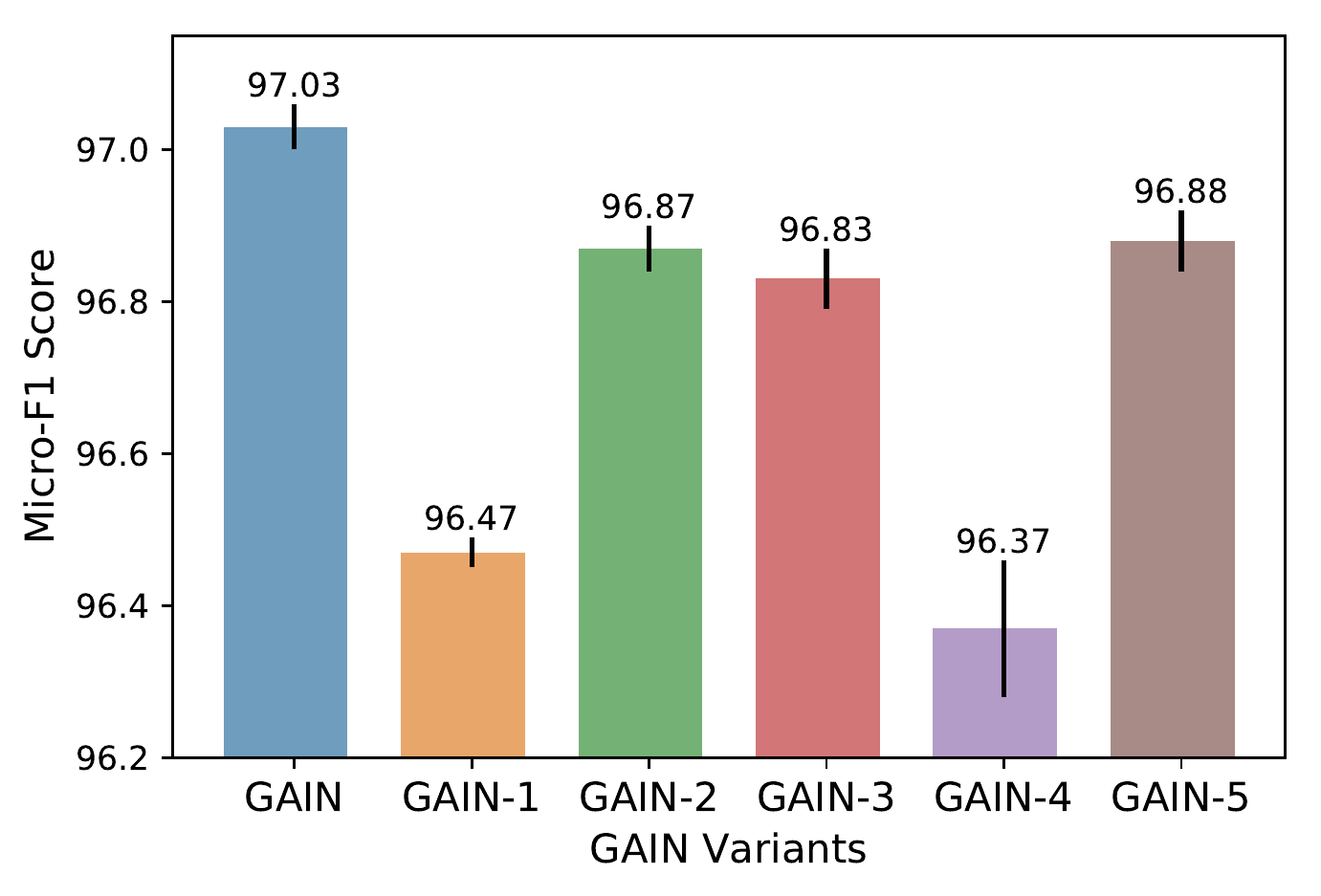}}
		\hspace*{3pt}
		\subfigure[Testing Micro-F1 on Pubmed dataset.] {\includegraphics[width=0.315\linewidth]{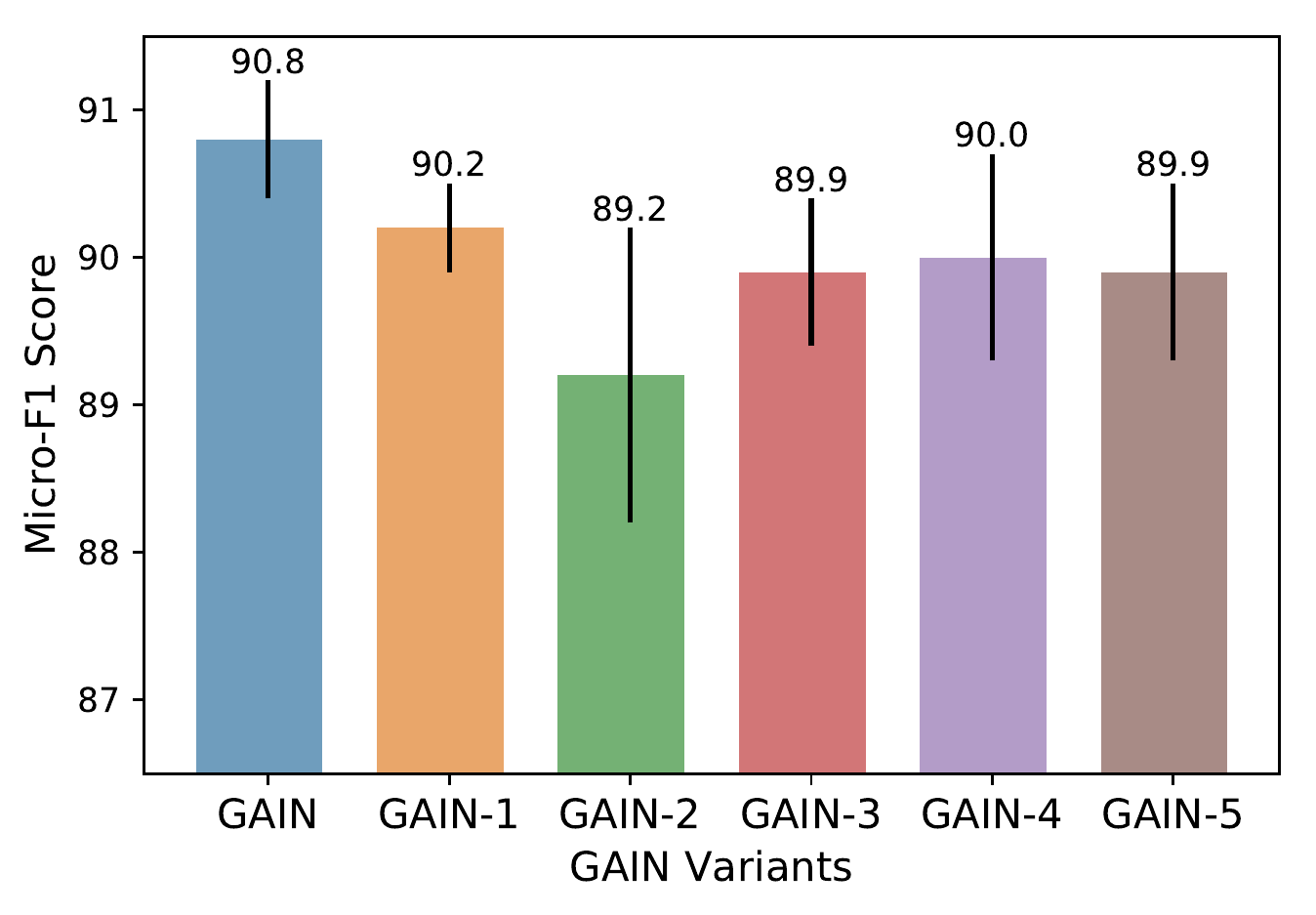}}
		
		\hspace*{10pt}
		\subfigure[Testing AUC on Financial News Dataset.] {\includegraphics[width=0.32\linewidth]{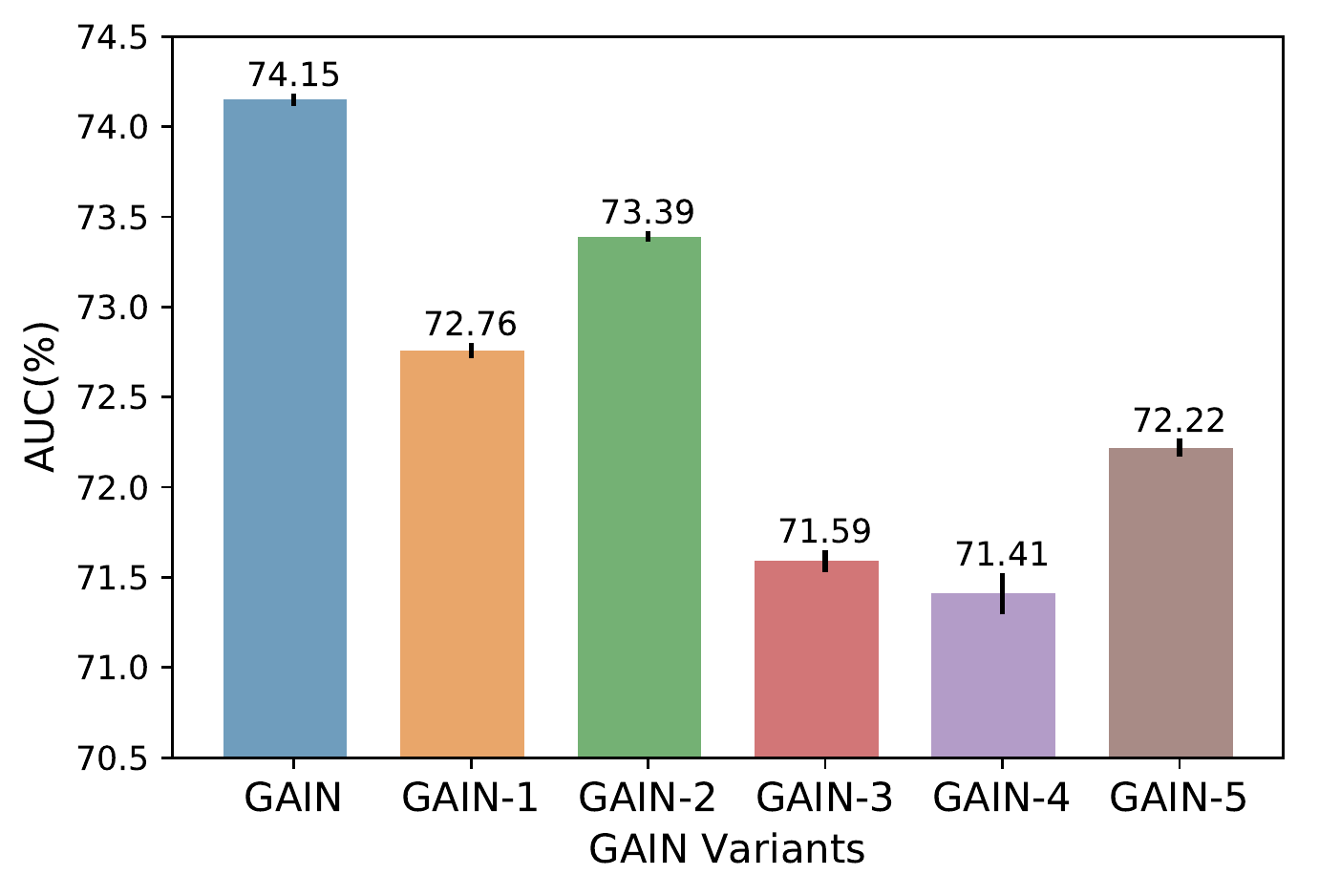}}
		\hspace*{10pt}
		\subfigure[Testing Logloss on Financial News Dataset.] {\includegraphics[width=0.32\linewidth]{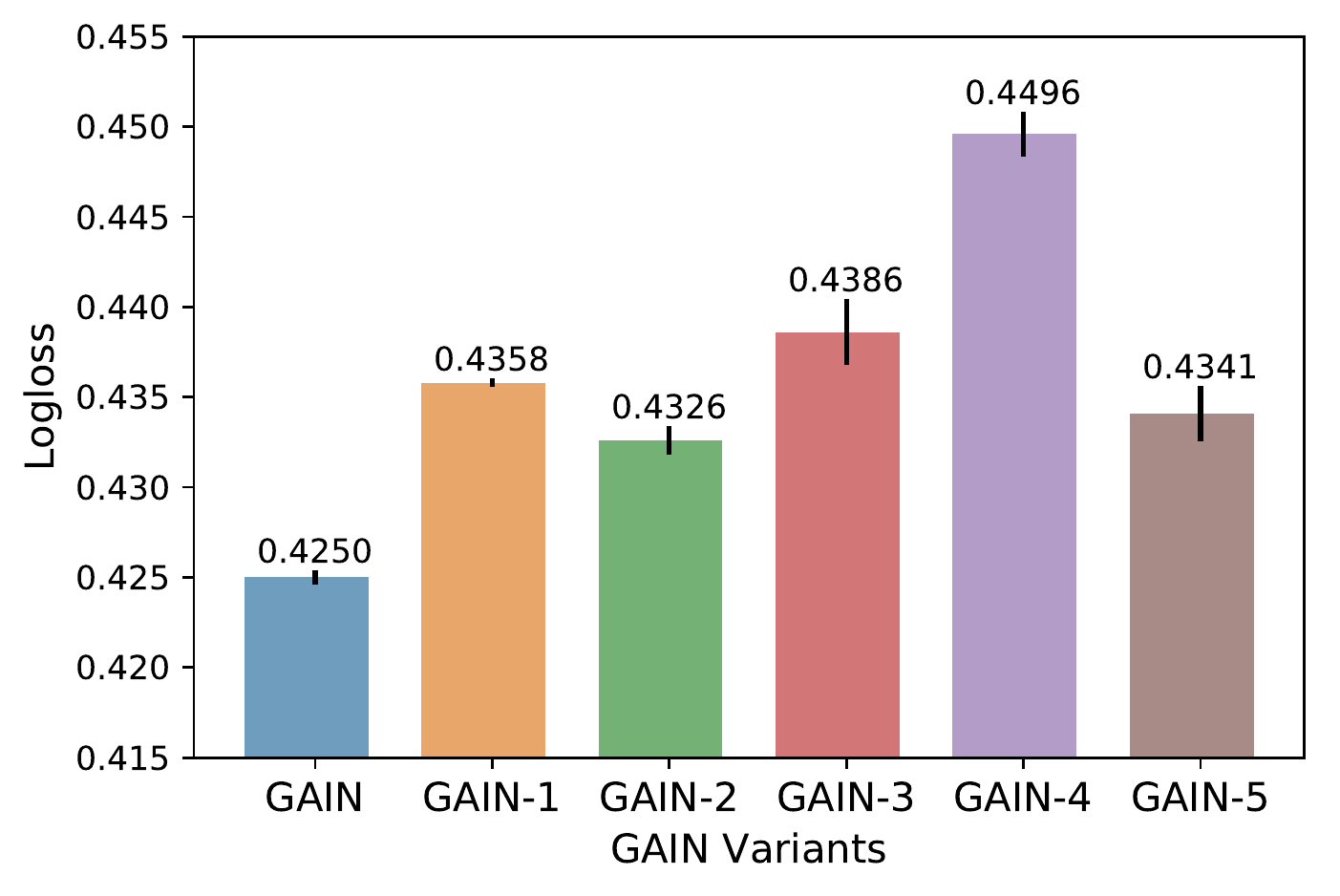}}
		
		\caption{Performance comparison of different GAIN variants.}
		\label{results-variants}
	\end{center}
	
\end{figure*}

Whether the user clicks on a news depends on the user's preferences and the content of the news. Therefore, in order to accurately predict the user's response to a news,  we need to take both the information of user and news into consideration. In this task, we compare our proposed GAIN model with GNNs with different aggregators. Besides, we also select several highly tuned models that are widely used in user click prediction problem for comparison, including FM(Factorization Machines)\cite{rendle2010factorization}, DeepFM\cite{guo2017deepfm}, DCN(Deep \& Cross Network) \cite{wang2017deep} and xDeepFM\cite{lian2018xdeepfm}. For these models, each sample contains two categorical fields:(1) news ID; (2) user ID. For a fair comparison, the news ID embedding matrix and user ID embedding matrix are respectively initialized by the user feature and news feature described in Section \label{data} like GNNs.

In this experiment, we use two common metrics in CTR prediction tasks for model evaluation: AUC (Area Under the ROC curve) and Logloss. AUC measures the probability that the model predicts higher score for a given true positive instance than a randomly selected negative one. One of the benefits of AUC is that it is insensitive to class imbalance problem. Logloss measures the distance between model's prediction and the ground truth and can be used as reference for ranking strategy. Therefore, these two metrics are widely used in CTR prediction problem.

From the results shown in Table \ref{news}, We can find that GNNs achieve high performance in the user response prediction task comparing with other baselines. By collecting neighborhood information on the heterogeneous network, GNNs can obtain information such as user preferences, and thus achieve better prediction results. Our proposed GAIN model outperforms the baseline models which demonstrates that GAIN is able to be applied to recommendation systems.

Besides, we explore the test performance of different GNN models with different neighborhood sample sizes. GraphSAGE-MEAN(HS) corresponds to the results that obtained by GraphSAGE with mean aggregator and heuristic sampling strategy. According to the results shown in Fig. \ref{sizes_fig}, we can draw our conclusion that our proposed GAIN model outperforms baselines continuously. In addition, when the sampling size is larger, the impact of heuristic sampling is reduced. This may be because most nodes can preserve all neighbor nodes with a large sampling size. This experiment demonstrates that GAIN can achieve a high performance on click-through rate prediction task with a small neighborhood sampling size.

\subsection{Effectiveness of Components}
As we have described above, the aggregation phase and update phase of our GAIN model are quite different from previous GNNs. In order to validate the effectiveness of each components of GAIN, we compare different variants of GAIN model by eliminating different key components systematically from GAIN.
These variants of GAIN are defined as:
\begin{itemize}
	\item GAIN: The complete GAIN model.
	\item GAIN-1: We remove the explicit feature interaction between center node's embedding and aggregated neighboring embedding to validate the effectiveness of this component.
	\item GAIN-2: In this variant, we evaluate the contribution of auto-encoder architecture used in aggregation phase. Therefore, it does not contain the auto-encoder module.
	\item GAIN-3: In order to verify the effectiveness of aggregator-level attention mechanism used in aggregation phase, we use single type of aggregator instead of multiple types of aggregators with attention mechanism in GAIN-3, GAIN-4 and GAIN-5. In GAIN-3, we only utilize average-pooling aggregator to capture the neighboring information.
	\item GAIN-4: In this variant, we only utilize max-pooling aggregator in aggregation phase.
	\item GAIN-5: In this variant, we only utilize importance pooling aggregator in aggregation phase.	
\end{itemize}

\begin{figure}[!h]
	
	\begin{center}
		
		\subfigure[Attention weights at hop-1.] {\includegraphics[width=1\linewidth]{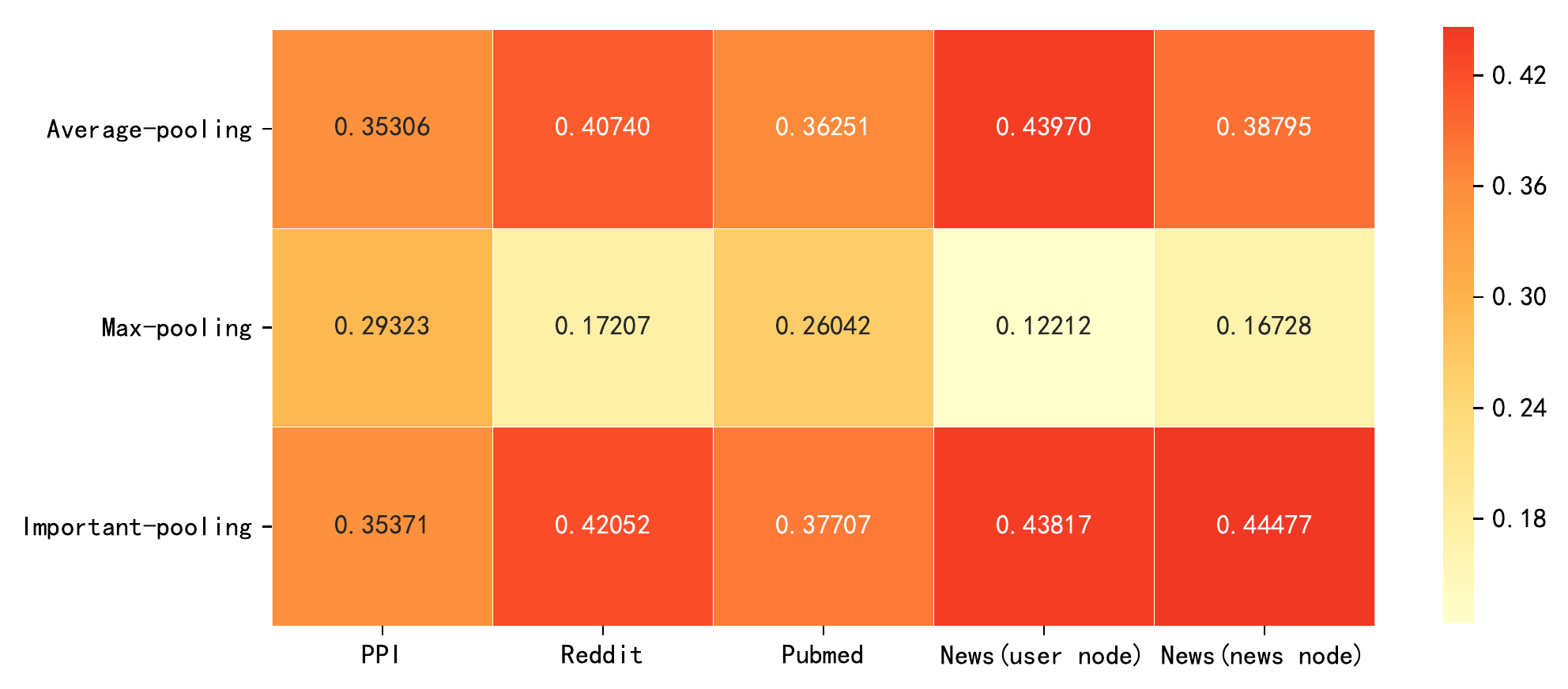}}
		\hspace*{2pt}
		\subfigure[Attention weights at hop-2.] {\includegraphics[width=1\linewidth]{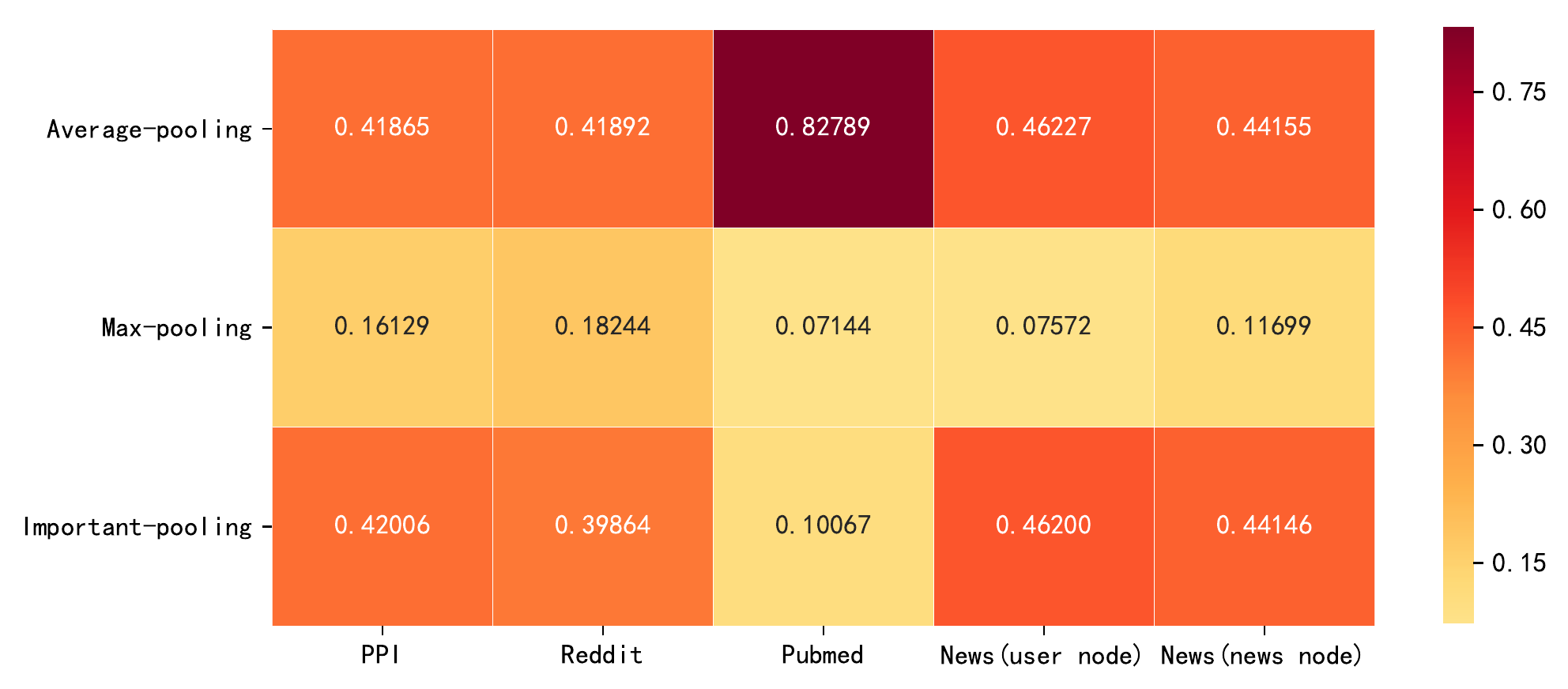}}
		
		\caption{Visualization of the average attention weights of different aggregators.}
		\label{heatmap}
	\end{center}
	
\end{figure}

The results are shown in Fig. \ref{results-variants}.  The complete GAIN model outperforms other variants and verify that each component has its contribution. GAIN-3, GAIN-4 and GAIN-5 validate that aggregator-level attention mechanism that utilizes different types of aggregators to capture neighboring information from different perspectives can boost the performance, especially on large datasets like Reddit and Financial News datasets. GAIN-1 and GAIN-2 also perform worse than GAIN, which suggests that explicit feature interactions operation and auto-encoder architecture can help improve the expression ability of the updated node embedding. Specifically, the graph feature interaction operations on the user-item graph can efficiently learn the cross information between user features and item features and help GAIN to capture user preferences.

In Fig.\ref{heatmap}, we visualization the average attention weights of the three types of aggregators on different tasks. Since the graph data of the financial news dataset contains news node and user node, Fig.\ref{heatmap} shows the average attention weight of each aggregator when the center node is news and user respectively. The Fig.\ref{heatmap} indicates that the same aggregator contributes differently at different depths from an average perspective. It should be noted that an aggregator with a large average attention weight may have little effect on some nodes since the weights of the same type of aggregator might be different for different center nodes.

The aggregation operation of GAT and GaAN uses a multi-head attention mechanism, and GaAN additionally introduces a convolutional network to calculate gate weights. In general, the parameter number of GAIN is on par with GNNs with multi-head attention mechanism such as GAT for the aggregation phase. Different from node-level attention mechanism used in GAT and GaAN which needs to calculate attention weight for every single neighbor node, the aggregator-level mechanism in GAIN calculates attention weights for aggregators, which reduce the computation.
Meanwhile, the aggregator-level attention mechanism is efficient since the aggregate operations of different aggregators are independent and parallelizable. Although GAIN introduces extra parameters and computations for the auto-encoder and feature interaction operations, GAIN can effectively learn the cross feature and graph structure information in the update phase. Due to effective aggregation and update operations, GAIN can achieve competitive performance with a small sampling neighborhood size.

\section{Conclusion}
In this paper, we study the problem of inductive learning on graphs. To address this challenging problem, we propose a novel graph neural network model, Graph Attention \& Interaction Network (GAIN). We introduce aggregator-level attention to the aggregation phase, which enables our GAIN model to aggregate neighboring information from different aspects explicitly. In the update phase, we exploit auto-encoder to learn the highly non-linear information of network and use explicit feature interactions to improve the expressing ability of updated node embeddings. To better preserve the graph structure information, we introduce the graph regularization to our loss function.
With comprehensive neighbors sampling scheme and mini-batch training fashion, our approach can be applied to larger-scale graphs and can handle unseen nodes or graphs.

We conduct extensive experiments on several benchmarks for inductive node classification and a financial news dataset for user response prediction. The experimental results confirm that our GAIN model evidently outperforms strong baselines including the state-of-the-art models.
%


%



%
\bibliographystyle{IEEEtran}
\bibliography{./ref}
\end{document}